\documentclass{article}
\usepackage{spconf,amsmath,graphicx}
\usepackage{amssymb}
\usepackage{pifont}
\usepackage{multicol, multirow}
\usepackage{url}

\usepackage{enumitem}
\setlist{nosep, leftmargin=14pt}
\usepackage[table]{xcolor} 
\usepackage{adjustbox}
\usepackage{mwe} 
\usepackage{wasysym}
\usepackage{float}

\title{From Specialist to Generalist: Unlocking SAM's Learning Potential on Unlabeled Medical Images}
\name{
\parbox{\textwidth}{\centering 
Vi Vu$^{1\star}$,
Thanh-Huy Nguyen$^{1\star}$,
Tien-Thinh Nguyen$^{2\star}$,
Ba-Thinh Lam$^{3}$ \\  
\textit{Hoang-Thien Nguyen}$^{1}$,
\textit{Tianyang Wang}$^{4}$,
\textit{Xingjian Li}$^{1\dagger}$,
\textit{Min Xu}$^{1\dagger}$
\thanks{$\dagger$ Corresponding Author: lixj04@gmail.com; mxu1@cs.cmu.edu} \thanks{$^{\ast}$ Equal contribution.}}
}
\address{
$^1$ Carnegie Mellon University, USA \\
$^2$ Industrial University of Ho Chi Minh City, Vietnam \\
$^3$ University of North Carolina at Charlotte, USA \\
$^4$ University of Alabama at Birmingham, USA 
}
\begin{document}
%
\maketitle
\begin{abstract}
Foundation models like the Segment Anything Model (SAM) show strong generalization, yet adapting them to medical images remains difficult due to domain shift, scarce labels, and the inability of Parameter-Efficient Fine-Tuning (PEFT) to exploit unlabeled data. While conventional models like U-Net excel in semi-supervised medical learning, their potential to assist a PEFT SAM has been largely overlooked. We introduce \textbf{SC-SAM}, a specialist-generalist framework where U-Net provides point-based prompts and pseudo-labels to guide SAM’s adaptation, while SAM serves as a powerful generalist supervisor to regularize U-Net. This reciprocal guidance forms a bidirectional co-training loop that allows both models to effectively exploit the unlabeled data. Across prostate MRI and polyp segmentation benchmarks, our method achieves state-of-the-art results, outperforming other existing semi-supervised SAM variants and even medical foundation models like MedSAM, highlighting the value of specialist–generalist cooperation for label-efficient medical image segmentation. Our code is available at \url{https://github.com/vnlvi2k3/SC-SAM}.
\end{abstract}

%


\section{Introduction}

Foundation models have reshaped image segmentation through large-scale pre-training, with the \textit{Segment Anything Model} (SAM) \cite{sam} showing impressive prompt-driven zero-shot performance. However, SAM struggles in medical imaging due to the significant domain gap, scarce annotations, and the instability of tuning a large model with limited labels. Parameter-Efficient Fine-Tuning (PEFT) reduces computation but still relies heavily on labeled data, leaving the abundant unlabeled medical images underutilized.

Meanwhile, conventional architectures such as U-Net have shown, over years of semi-supervised research, that unlabeled data can be a powerful asset. Techniques like consistency regularization and pseudo-labeling allow these models to extract reliable structures from unannotated images. Interestingly, although U-Nets excel at learning from unlabeled data, SAM does not naturally inherit this ability. SemiSAM+ \cite{semisam+} prompts a frozen SAM using a specialist. KnowSAM \cite{knowsam} distills SAM into lighter networks. CPC-SAM \cite{cpcsam} enforces cross-prompt consistency between dual SAM decoders. These methods demonstrate various uses of SAM but also introduce architectural complexity or rely solely on SAM’s predictions. It leaves open the question of whether a conventional model could meaningfully help a PEFT SAM learn from unlabeled data.

This observation motivates our approach. We propose \textit{SC-SAM: Specialist Co-training with SAM}, a specialist-generalist collaboration framework that leverages the complementary strengths of U-Net and SAM. Instead of treating SAM as a primary supervisor, we invert the direction: U-Net first learns from both labeled and unlabeled data using standard semi-supervised strategies, and its predictions are then transformed into point-based prompts and pseudo-labels that guide SAM during PEFT. In parallel, SAM generates refined masks that regularize and stabilize U-Net’s training. This forms a \emph{bidirectional co-training loop} in which U-Net contributes structural domain knowledge, and SAM serves as a high-level semantic regularizer. The two models progressively align, allowing SAM to benefit directly from unlabeled information.

Across various medical image segmentation benchmarks, \textit{SC-SAM} achieves consistent improvements over existing semi-supervised SAM variants and even surpasses strong medical foundation models such as MedSAM and SAM-Med-2D. Our results point to a simple but underexplored principle: Cooperation between conventional specialists and modern generalist models can unlock new pathways for scalable, label-efficient medical image segmentation.


\begin{figure*}[t]
  \centering
  \includegraphics[width=0.9\linewidth]{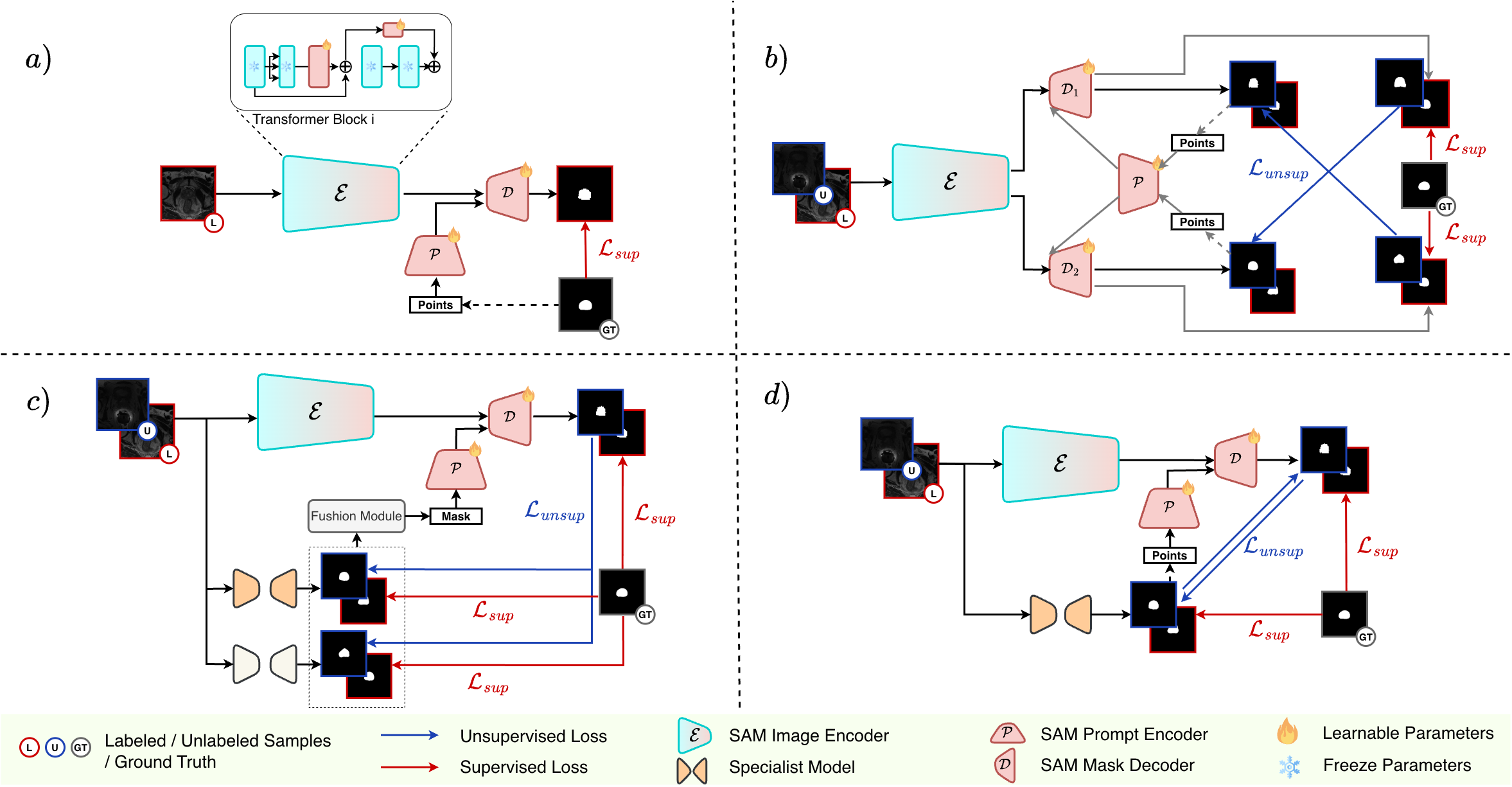}
  \caption{Overview of different techniques for incorporating SAM in semi-supervised settings: a) PEFT-SAM, b) Dual-SAM, c) SP-SAM, and d) SC-SAM (Ours)}
  \label{fig:framework}
\end{figure*}

\section{Methodology}

In semi-supervised segmentation, the training set consists of a labeled set $D_l = \{(X_i^l, Y_i^l)\}_{i=1}^{N}$, and a larger unlabeled set $D_u = \{(X_j^u)\}_{j=1}^{M}$ ($M \gg N$). Each image $X \in \mathbb{R}^{C \times H \times W}$ has $C=3$ for RGB or $C=1$ for grayscale inputs, and each labeled sample $X_i^l$ is paired with a ground-truth mask $Y_i^l$. We next outline existing techniques for incorporating SAM in semi-supervised settings: PEFT-SAM, Dual-SAM, and SP-SAM, and show how their limitations give rise to our method, SC-SAM.

\label{sec:format}

\noindent \textbf{PEFT-SAM: Parameter-efficient Fine-tuning of SAM on labeled set.}
To adapt SAM on the limited labeled data, we apply PEFT  by introducing learnable Adapter layers \cite{samadapter} into each Transformer block of its image encoder $\mathcal{E}$, while prompt encoder $\mathcal{P}$ and mask decoder $\mathcal{D}$ are kept trainable with negligible additional parameters as outlined in Fig.~\ref{fig:framework}(a). Point prompts are sampled from the ground-truth masks of the labeled set and, together with the encoder's image embeddings, are fed into the decoder to obtain the predictions:
\begin{equation*}
\text{points} = \mathrm{Sample}(Y^{l})
\end{equation*}
\begin{equation*}
P_{SAM}^{l} = \mathcal{D}\big(\mathcal{E}(X^{l}), \mathcal{P}(\text{points})\big),
\end{equation*}
where $\mathrm{Sample}(\cdot)$ randomly selects five foreground and five background points from the ground truth. The supervised loss  $\mathcal{L}_{\mathrm{seg}}$ is the mean of Dice and Cross-Entropy losses:

\begin{equation*}
\mathcal{L}_{\text{sup}} = \mathcal{L}_{seg}\big(\mathrm{softmax}(P_{SAM}^{l}), Y^{l}\big)
\end{equation*}


\noindent \textbf{Dual-SAM: Leveraging unlabeled data with Dual-branch SAM.} Miao \textit{et al.} \cite{cpcsam} introduced a dual-decoder SAM framework, where each decoder generates prompts and provides cross-supervision to the other. We denote $\mathcal{D}_1$ and $\mathcal{D}_2$ as two distinct learnable decoders (see Fig.~\ref{fig:framework}~(b)).

\noindent \textbf{Phase 1:}
\begin{equation*}
P^{l,u}_{1,2} = \mathcal{D}_{1,2}\big(\mathcal{E}(X^{u,l})\big)
\end{equation*}
\begin{equation*}
\text{points}_{1,2} = \mathrm{Sample}\big(P^{l,u}_{1,2}\big)
\end{equation*}
\textbf{Phase 2:}
\begin{equation*}
P^{l,u}_{1} = \mathcal{D}_{1}\big(\mathcal{E}(X^{u,l}), \text{points}_{2}\big),\quad
P^{l,u}_{2} = \mathcal{D}_{2}\big(\mathcal{E}(X^{u,l}), \text{points}_{1}\big)
\end{equation*}
The pseudo-label $\hat{P} = \arg\max(P)$ is obtained from the model prediction for computing the unsupervised loss:
\begin{equation*}
\mathcal{L}_{\text{sup}} = \mathcal{L}_{seg}(P^{l}_{1}, Y^{l})
                         + \mathcal{L}_{seg}(P^{l}_{2}, Y^{l})
\end{equation*}
\begin{equation*}
\mathcal{L}_{\text{unsup}} = \mathcal{L}_{seg}(P^{u}_{1}, \hat{P}^{u}_{2})
                           + \mathcal{L}_{seg}(P^{u}_{2}, \hat{P}^{u}_{1})
\end{equation*}


\noindent \textbf{SP-SAM: Specialist-guided Prompting for SAM fine-tuning.} Fig. \ref{fig:framework} (c) represents the work of Huang \textit{et al.} \cite{knowsam}, which integrates two specialists (or conventional networks) UNet \cite{unet} $\mathcal{S}_1$ and VNet \cite{vnet} $\mathcal{S}_2$, together with a learnable fusion module $\mathcal{F}$ that fuses their predicted masks to generate mask prompts for SAM: 
\begin{equation*}
P^{l,u}_{1,2} = S_{1,2}(X^{l,u})
\end{equation*}
\begin{equation*}
\text{mask} = \mathcal{F}\big(P^{l,u}_{1}, P^{l,u}_{2}\big)
\end{equation*}
\begin{equation*}
P^{l,u}_{\text{SAM}} = \mathcal{D}\big(\mathcal{E}(X^{l,u}), \mathcal{P}(\text{mask})\big)
\end{equation*}
Through knowledge distillation, SAM supervises the specialists and allows them to exploit the unlabeled set to produce more robust prompts for SAM fine-tuning. 
\begin{equation*}
\mathcal{L}_{\text{sup}} = \mathcal{L}_{seg}(P^{l}_{1}, Y^{l}) 
                         + \mathcal{L}_{seg}(P^{l}_{2}, Y^{l}) 
                         + \mathcal{L}_{seg}(P^{l}_{\text{SAM}}, Y^{l})
\end{equation*}
\begin{equation*}
\mathcal{L}_{\text{unsup}} = \mathcal{L}_{KD}(P^{u}_{1}, \hat{P}^{u}_{\text{SAM}}) 
                           + \mathcal{L}_{KD}(P^{u}_{2}, \hat{P}^{u}_{\text{SAM}}),
\end{equation*}
where $L_{\text{KD}}$ is the Kullback–Leibler divergence loss.
    
\noindent \textbf{SC-SAM: Specialist Co-training with SAM.} \textit{PEFT-SAM} focuses on fine-tuning SAM using limited labeled data or adopt zero or few-shot tuning, leaving the unlabeled set under-exploited. \textit{Dual-SAM} leverages unlabeled data through dual-branch cross-supervision; however, under significant domain shift in semi-supervised medical image segmentation, generalist models like SAM are prone to overconfident predictions, leading both branches to step into coupling problem and converge to similar local minima. In contrast, specialist networks (e.g., UNet) trained from scratch can correct SAM’s high-confidence false predictions. Unlike \textit{SP-SAM}, which exclusively exploits specialist strengths only for robust prompting, our goal is to collaboratively train generalist and specialist networks to form a bi-directional co-training loop (see Fig.~\ref{fig:framework}~(d)).
\begin{equation}
P_{\text{UNet}}^{l,u} = \mathcal{S}(X^{l,u})
\end{equation}
\begin{equation}
\text{points} = \text{Sample}(P_{\text{UNet}}^{l,u})
\end{equation}
\begin{equation}
P_{\text{SAM}}^{l,u} = \mathcal{D}\big(\mathcal{E}(X^{l}), \mathcal{P}(\text{points})\big)
\end{equation}
However, naive co-training is unstable due to the disparity in convergence behaviors: SAM shows superior early-stage performance to serve as an effective regularizer for UNet but also suffer from early overfitting, while UNet converges slowly yet ultimately adapts better to the target domain. Directly transferring supervision signals from UNet to SAM causes noisy unlabeled signals dominate in the early unsupervised phase and corrupt SAM. We thereby design a sigmoid ramp-up strategy as follows:
\begin{equation}
\mathcal{L}_{\text{sup}} = 
\mathcal{L}_{\text{seg}}\!\left(P_{\text{UNet}}^{l}, Y^{l}\right) +
\mathcal{L}_{\text{seg}}\!\left(P_{\text{SAM}}^{l}, Y^{l}\right)
\end{equation}
\begin{equation}
\mathcal{L}_{\text{unsup}} = 
\mathcal{L}_{\text{seg}}\!\left(P_{\text{UNet}}^{u}, \hat{P}_{\text{SAM}}^{u}\right) +
\omega(t)\, \mathcal{L}_{\text{seg}}\!\left(P_{\text{SAM}}^{u}, \hat{P}_{\text{UNet}}^{u}\right)
\end{equation}
\begin{equation}
\omega(t) =
\begin{cases}
e^{-\left(1 - \frac{t}{T_{\max}}\right)^2}, & 0 \le t \le T_{\max}, \\
1, & t > T_{\max}
\end{cases}
\end{equation}
where $t$ and $T_{\max}$ are the current iteration and the ramp up length, respectively. The final training objective is:
\begin{equation}
\mathcal{L}_{\text{total}} = \mathcal{L}_{\text{sup}} + \mathcal{L}_{\text{unsup}}
\end{equation}

\section{Experiment results}

All experiments were performed on a single NVIDIA RTX 3090 Ti GPU (24 GB RAM) using Python 3.10, PyTorch 2.9, and CUDA 12.8. We used Adam optimizer (initial learning rate $10^{-4}$) for SAM and SGD (momentum 0.9) for UNet. The batch size was set to 24, consisting of 12 labeled and 12 unlabeled samples. For labeled samples, we applied a weak augmentation pipeline including random flipping, brightness and contrast adjustments, shift-scale-rotate, and dropout. For unlabeled samples, we additionally used stronger augmentations such as grid distortion and shift-scale-rotate. For \textit{PEFT-SAM} and \textit{SC-SAM}, we also employed learnable bounding box prompts following \textit{SP-SAM} \cite{knowsam}. 
\subsection{Datasets}
We evaluated our method on two public medical image segmentation benchmarks. The \textbf{PROMISE12} \cite{promise} dataset consists of 50 prostate MR volumes, split into 35/5/10 for train/val/test. Slices within each MRI volume were treated as 2D images. For the \textbf{COLON} dataset, we used 612 endoscopic images from CVC-ClinicDB \cite{cvcclinicdb} and 838 images from Kvasir \cite{kvasir}, and evaluated the cross-dataset generalization on five benchmark test sets: CVC-300 \cite{cvc300}, CVC-ColonDB \cite{cvccolondb}, CVC-ClinicDB, ETIS-LaribPolypDB \cite{etis}, and Kvasir-SEG \cite{kvasir}. For both datasets, only 5\% or 10\% of the training images were provided with annotations.

\subsection{Quantitative results}

\begin{table}[htbp]
\vspace{-5mm}
\centering
\caption{Quantitative results on the \textbf{PROMISE12} dataset.\\
}
\label{tab:promise_quan}
\setlength{\tabcolsep}{2pt}
\renewcommand{\arraystretch}{1.15}
\scriptsize 
\begin{tabular}{c|l|cccc|cccc}
\hline
\multirow{2}{*}{\shortstack{\textbf{Spec.}\\\textbf{Avail}}}
 & \textbf{Labeled ratio} &
\multicolumn{4}{c|}{\textbf{5\%}} &
\multicolumn{4}{c}{\textbf{10\%}} \\
\cline{2-10}
 & \textbf{Model} & \textbf{Dice} & \textbf{IoU}  & \textbf{HD95}  & \textbf{ASD}
 & \textbf{Dice}  & \textbf{IoU}  & \textbf{HD95}  & \textbf{ASD}  \\
\hline
\multirow{4}{*}{\ding{55}} 
 & SAM \cite{sam} & 72.26 & 59.39 & 4.25 & 8.79 & 79.66 & 68.53 & 4.02 & 5.79 \\
 & MedSAM \cite{medsam} & 63.00 & 48.65 & 4.85 & 14.91 & 74.75 & 62.12 & 4.28 & 7.45 \\
 & SAM-Med2D \cite{sammed2d} & 49.18 & 35.85 & 10.74 & 15.58 & 62.78 & 48.49 & 4.51 & 9.29 \\
 & CPC-SAM \cite{cpcsam} & 73.73 & 62.48  & 26.49 & 10.22 & 80.42 & 70.75  & 12.04 & 5.99  \\
\hline
\multirow{2}{*}{$\checkmark$} 
 & KnowSAM \cite{knowsam} & \underline{78.49} & \underline{67.16} & \underline{4.11} & \underline{6.40} & \underline{82.93} & \underline{73.18} & \underline{3.94} & \underline{3.92} \\

 & \textbf{Ours} & \cellcolor{red!7} \textbf{83.64} & \cellcolor{red!7} \textbf{73.87} & \cellcolor{red!7} \textbf{3.98} & \cellcolor{red!7} \textbf{3.79} & \cellcolor{red!7} \textbf{83.35} & \cellcolor{red!7} \textbf{73.54} & \cellcolor{red!7} \textbf{3.91} & \cellcolor{red!7} \textbf{3.81} \\
\hline
\end{tabular}
\end{table}

\begin{table*}[htbp]
\centering
\caption{Quantitative results of different methods on the \textbf{Colon} datasets on 5\% and 10\% labeled fractions.}
\label{tab:colon_quan}
\vskip.2cm
\setlength{\tabcolsep}{4pt}
\renewcommand{\arraystretch}{1.15}
\scriptsize
\begin{tabular}{c|c|l|cc|cc|cc|cc|cc}
\hline
\multirow{2}{*}{\shortstack{\textbf{Labeled}\\\textbf{ratio}}} & \multirow{2}{*}{\shortstack{\textbf{Spec.}\\\textbf{Avail}}} & \textbf{Dataset} &
\multicolumn{2}{c|}{\textbf{CVC-300}} &
\multicolumn{2}{c|}{\textbf{CVC-ClinicDB}} &
\multicolumn{2}{c|}{\textbf{CVC-ColonDB}} &
\multicolumn{2}{c|}{\textbf{ETIS-Larib}} &
\multicolumn{2}{c}{\textbf{Kvasir}} \\
\cline{3-13}
 & & \textbf{Model} & Dice ($\uparrow$) & IoU ($\uparrow$) & Dice ($\uparrow$) & IoU ($\uparrow$) & Dice ($\uparrow$) & IoU ($\uparrow$) & Dice ($\uparrow$) & IoU ($\uparrow$) & Dice ($\uparrow$) & IoU ($\uparrow$) \\
\hline
\multirow{7}{*}{\textbf{5\%}}
 & \multirow{5}{*}{\ding{55}}
 & SAM \cite{sam} & 82.02 & 73.13 & 74.27 & 65.93 & 60.82 & 52.37 & 51.09 & 42.68 & 82.79 & 75.09 \\
 &  & MedSAM \cite{medsam} & 17.56 & 12.36 & 49.97 & 41.39 & 16.85 & 11.57 & 12.38 & 15.65 & 54.88 & 42.68 \\
 &  & SAM-Med2D \cite{sammed2d} & 14.09 & 8.10 & 41.73 & 31.22 & 18.73 & 11.45 & 13.09 & 7.70 & 38.73 & 26.23 \\
 &  & CPC-SAM \cite{cpcsam} & 55.56 & 50.52 & 61.05 & 54.73 & 44.68 & 40.17 & 37.62 & 34.11 & 80.87 & 74.26\\
 \cline{2-13}
 & \multirow{2}{*}{$\checkmark$}
 & KnowSAM \cite{knowsam} & \underline{83.75} & \underline{75.2} & \underline{78.86} & \underline{70.93} & \underline{66.76} & \underline{58.14} & \underline{56.72} & \underline{47.36} & \underline{84.78} & \underline{77.03} \\
  &  & \textbf{Ours} & \cellcolor{red!7} \textbf{88.72} & \cellcolor{red!7} \textbf{80.56} & \cellcolor{red!7} \textbf{79.54} & \cellcolor{red!7} \textbf{73.21} & \cellcolor{red!7} \textbf{67.77} & \cellcolor{red!7} \textbf{58.97} & \cellcolor{red!7} \textbf{56.91} & \cellcolor{red!7} \textbf{49.85} & \cellcolor{red!7} \textbf{84.79} & \cellcolor{red!7} \textbf{77.45} \\
\hline
\multirow{7}{*}{\textbf{10\%}}
 & \multirow{5}{*}{\ding{55}}
 & SAM \cite{sam} & 83.33 & 74.89 & 79.64 & 72.46 & 66.74 & 58.17 & 57.46 & 49.5 & 85.54 & 78.25 \\
 & & MedSAM \cite{medsam} & 27.42 & 20.82 & 59.33 & 49.79 & 22.18 & 15.89 & 23.45 & 17.18 & 57.42 & 45.29 \\
 & & SAM-Med2D \cite{sammed2d} & 13.59 & 7.78 & 49.29 & 36.9 & 17.63 & 10.8 & 15.44 & 9.39 & 41.66 & 28.51 \\
 & & CPC-SAM \cite{cpcsam} & 75.66 & 69.05 & 57.8 & 51.92 &  47.53 & 43.01 & 33.53 & 31.67 & 79.78 & 73.59 \\
 \cline{2-13}
 & \multirow{2}{*}{$\checkmark$}
 & KnowSAM \cite{knowsam} & \underline{86.49} & \underline{78.65} & \underline{82.54} & \underline{75.26} & \underline{68.21} & \underline{60.04} & \cellcolor{red!7} \textbf{64.11} & \cellcolor{red!7} \textbf{55.46} & \cellcolor{red!7} \textbf{86.64} & \cellcolor{red!7} \textbf{79.58} \\
 &  & \textbf{Ours} & \cellcolor{red!7} \textbf{88.17} & \cellcolor{red!7} \textbf{80.79} & \cellcolor{red!7} \textbf{83.06} & \cellcolor{red!7} \textbf{77.05} & \cellcolor{red!7} \textbf{68.69} & \cellcolor{red!7} \textbf{60.60} & \underline{60.07} & \underline{52.97} & \underline{86.27} & \underline{79.52} \\
\hline
\end{tabular}
\end{table*}

We report Dice coefficient, Intersection-over-Union (IoU), Average Surface Distance (ASD), and the 95th percentile Hausdorff Distance (HD95) for quantitative evaluation. Our proposed \textit{SC-SAM} was compared with several state-of-the-art baselines: For \textit{PEFT-SAM}, we adopted SAM \cite{sam}, SAM-Med2D \cite{sammed2d}, MedSAM \cite{medsam} as backbones; we used KnowSAM \cite{knowsam} to represent \textit{SP-SAM}, and CPC-SAM \cite{cpcsam} for \textit{Dual-SAM}. As shown in Tab. \ref{tab:promise_quan} and \ref{tab:colon_quan}, our method consistently achieves superior performance when trained with 5\% and 10\% labeled settings on PROMISE12 and COLON datasets. Especially at the 5\% labeled ratio, our \textit{SC-SAM} outperforms KnowSAM by a large margin, which ranks second in nearly all settings, highlighting that, in addition to using the specialists for prompt creation, effectively leveraging them to generate pseudo-labels for potential unlabeled data is critical. For \textit{PEFT-SAM} methods, SAM achieves higher accuracy than the medical SAM models, highlighting its strong generalizability when fine-tuned on limited data of the target domain. CPC-SAM performs relatively poorly, suggesting that leveraging UNet as a supervisor for unlabeled data eliminates its over-confidence and coupling issues more effectively than using another SAM.

\subsection{Qualitative results}

\setlength{\tabcolsep}{1pt} 
\renewcommand{\arraystretch}{0.5}

\begin{center}
\vspace{-5mm}

\begin{table}[htbp]
\caption{Qualitative results of different methods, {\color{red} $\Circle$} and {\color{green} $\CIRCLE$} denote groundtruth and predicted mask, respectively.\\
}
\label{tab:qual}
\begin{tabular}{c c c c c c c}
    & \textbf{\tiny SAM} & \textbf{\tiny MedSAM} & \textbf{\tiny SAM-Med2D} & \textbf{\tiny CPC-SAM} & \textbf{\tiny KnowSAM} & \textbf{\tiny Ours} \\

    \adjustbox{valign=c,rotate=90}{\tiny \textbf{5\% Promise}} &
    \includegraphics[width=0.15\columnwidth]{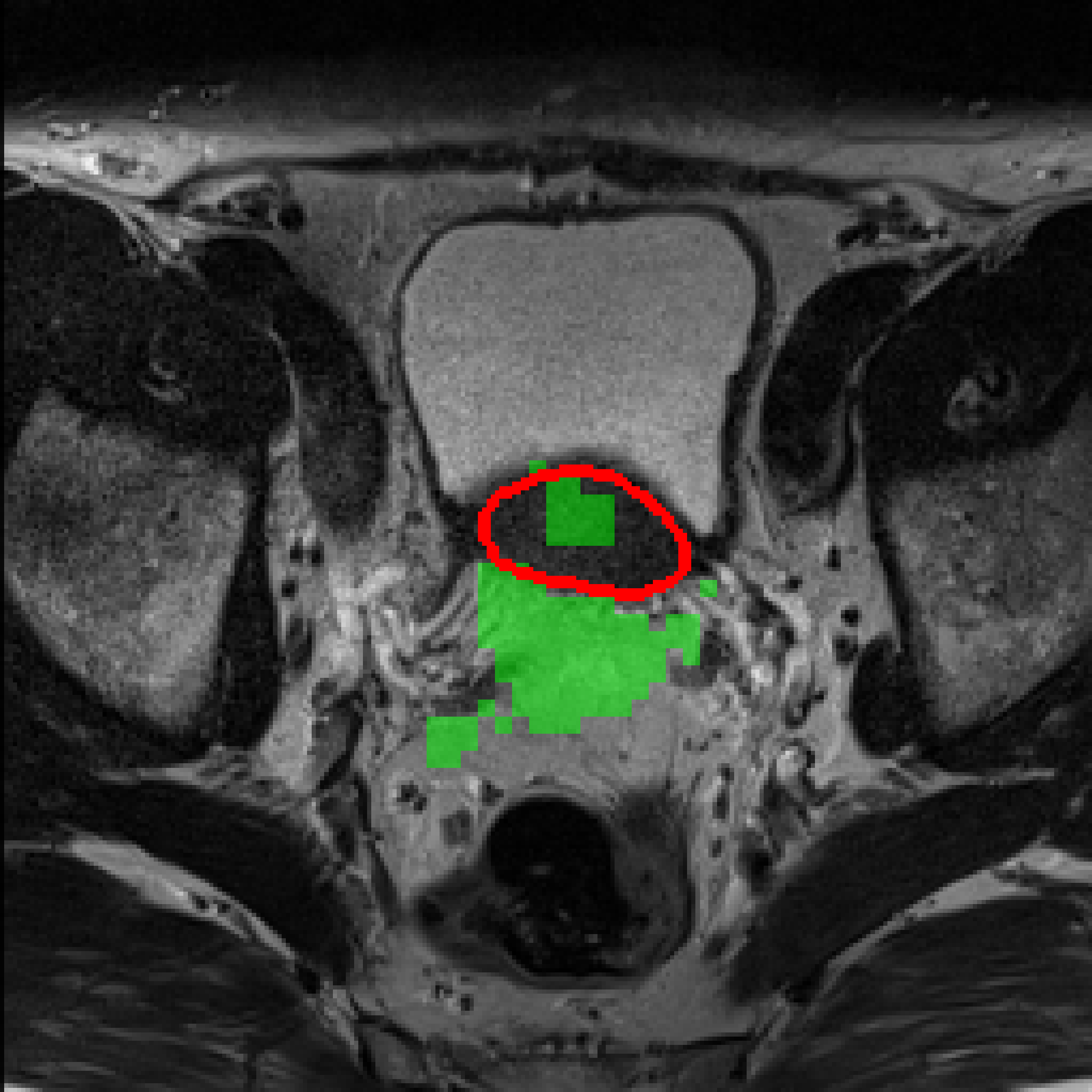} & 
    \includegraphics[width=0.15\columnwidth]{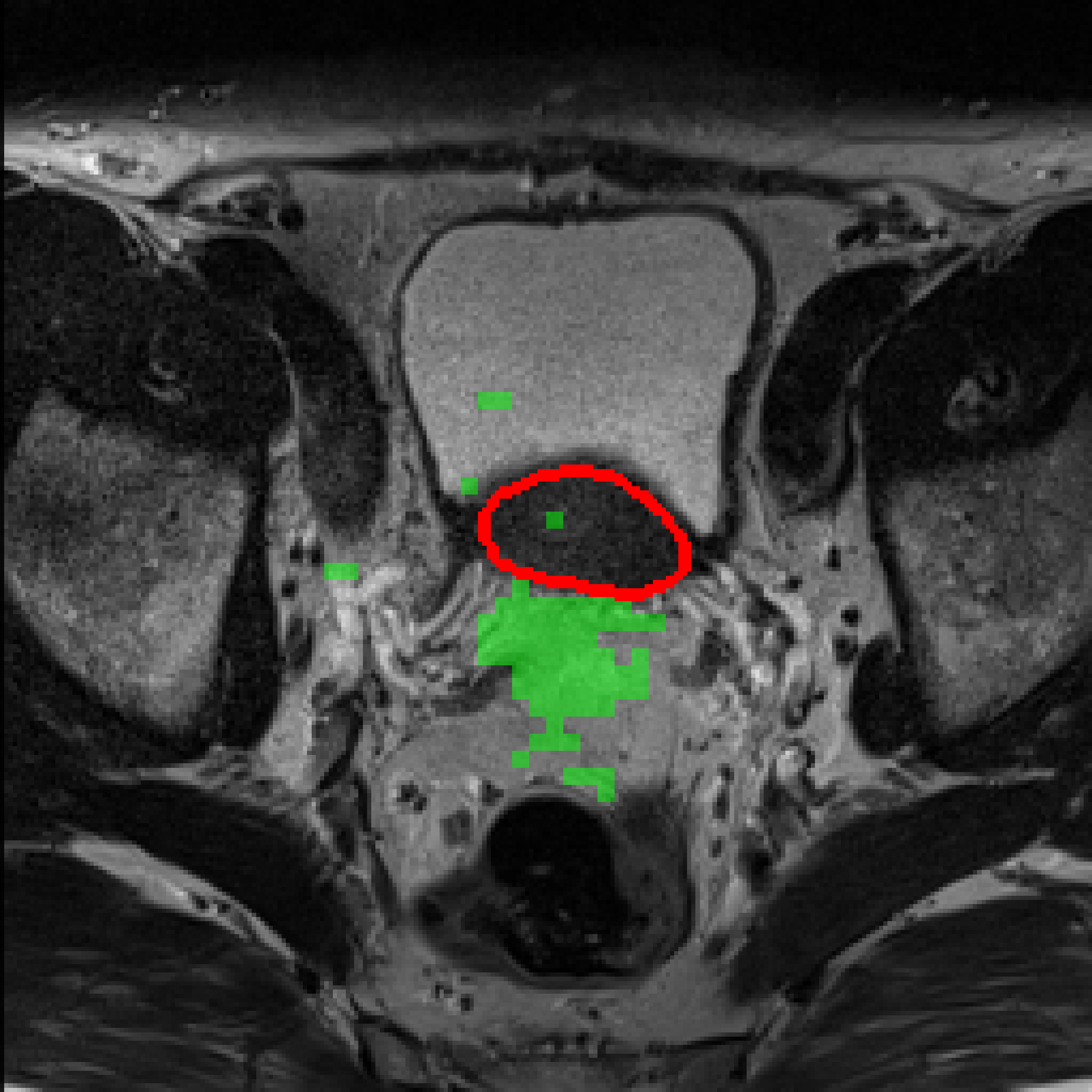} &
    \includegraphics[width=0.15\columnwidth]{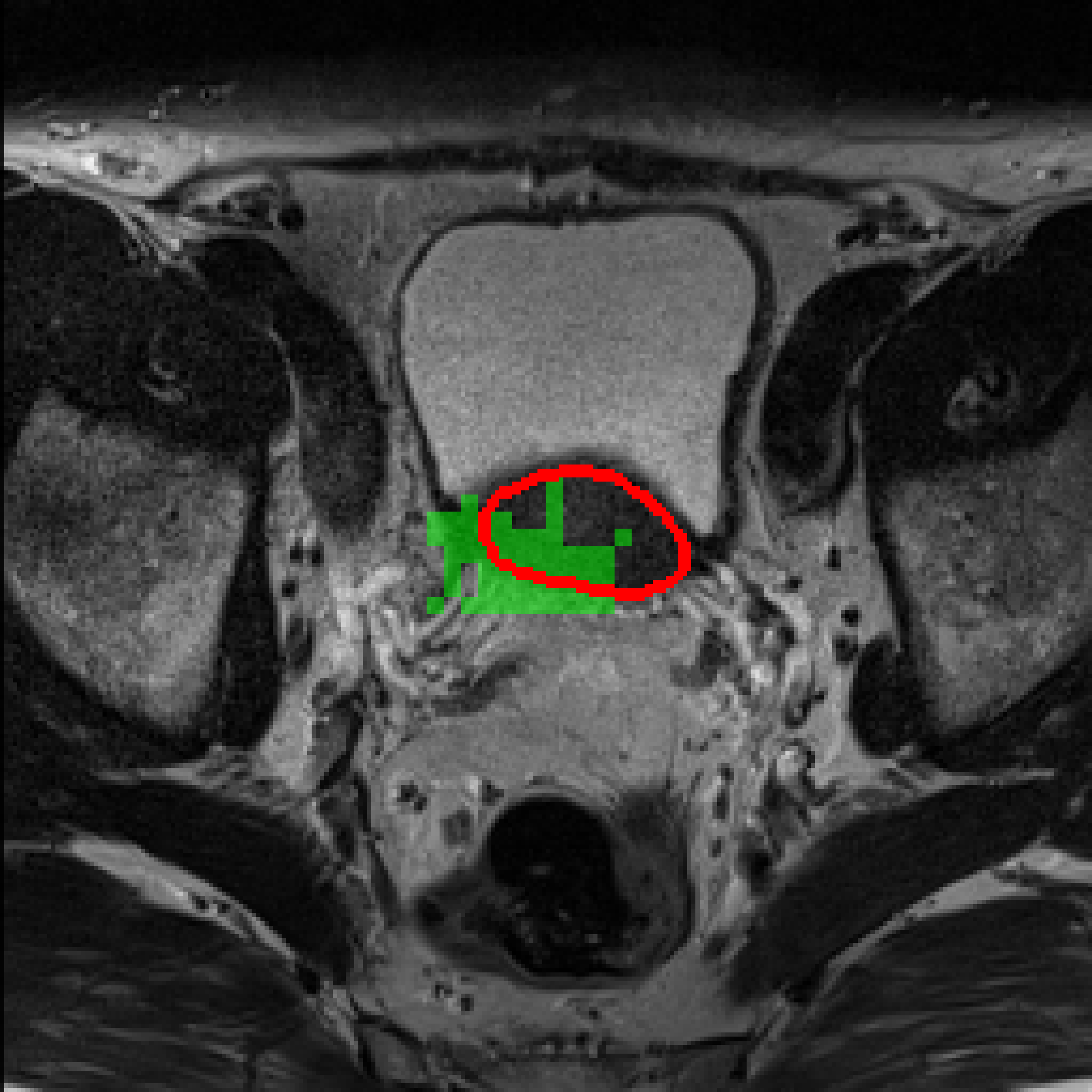} &
    \includegraphics[width=0.15\columnwidth]{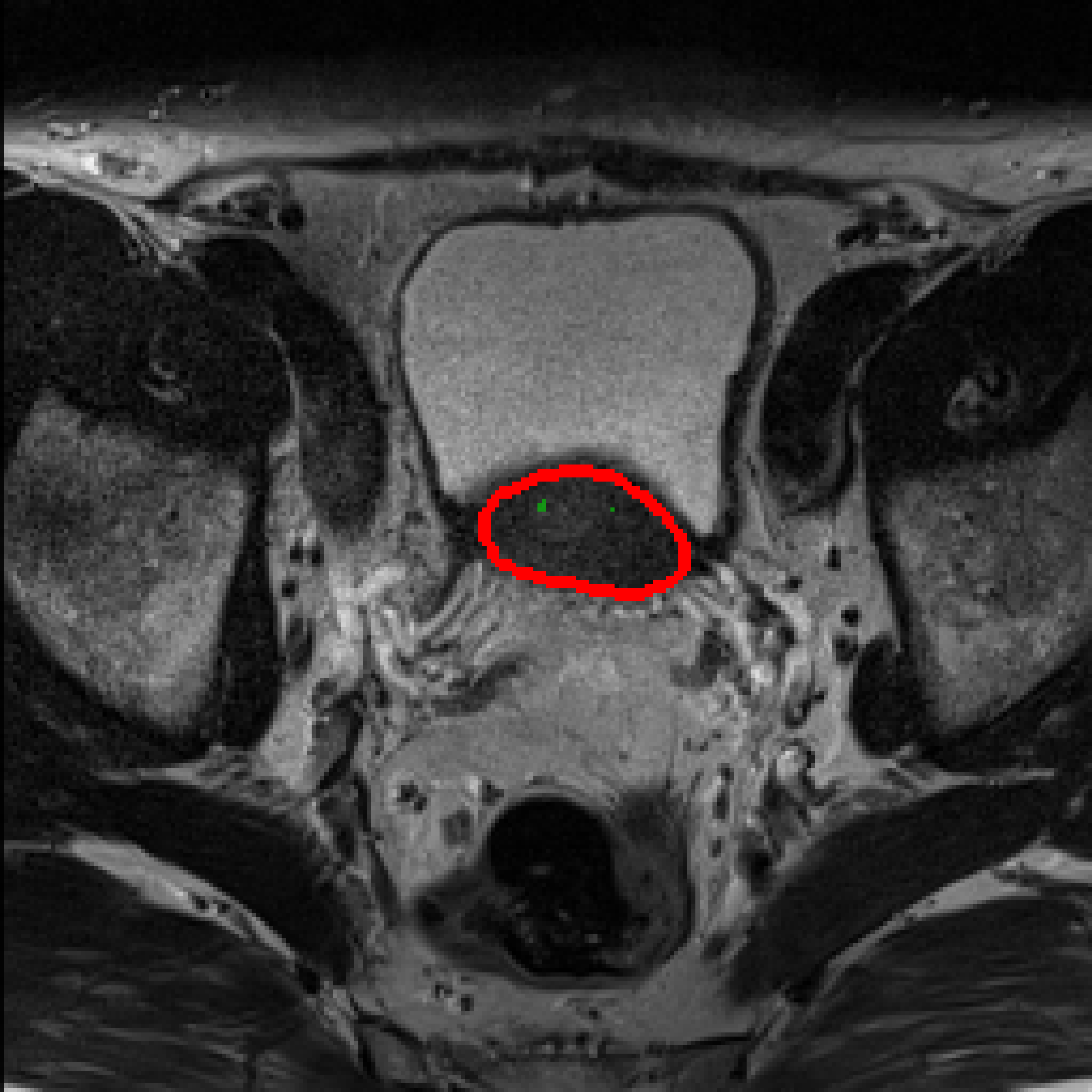} &
    \includegraphics[width=0.15\columnwidth]{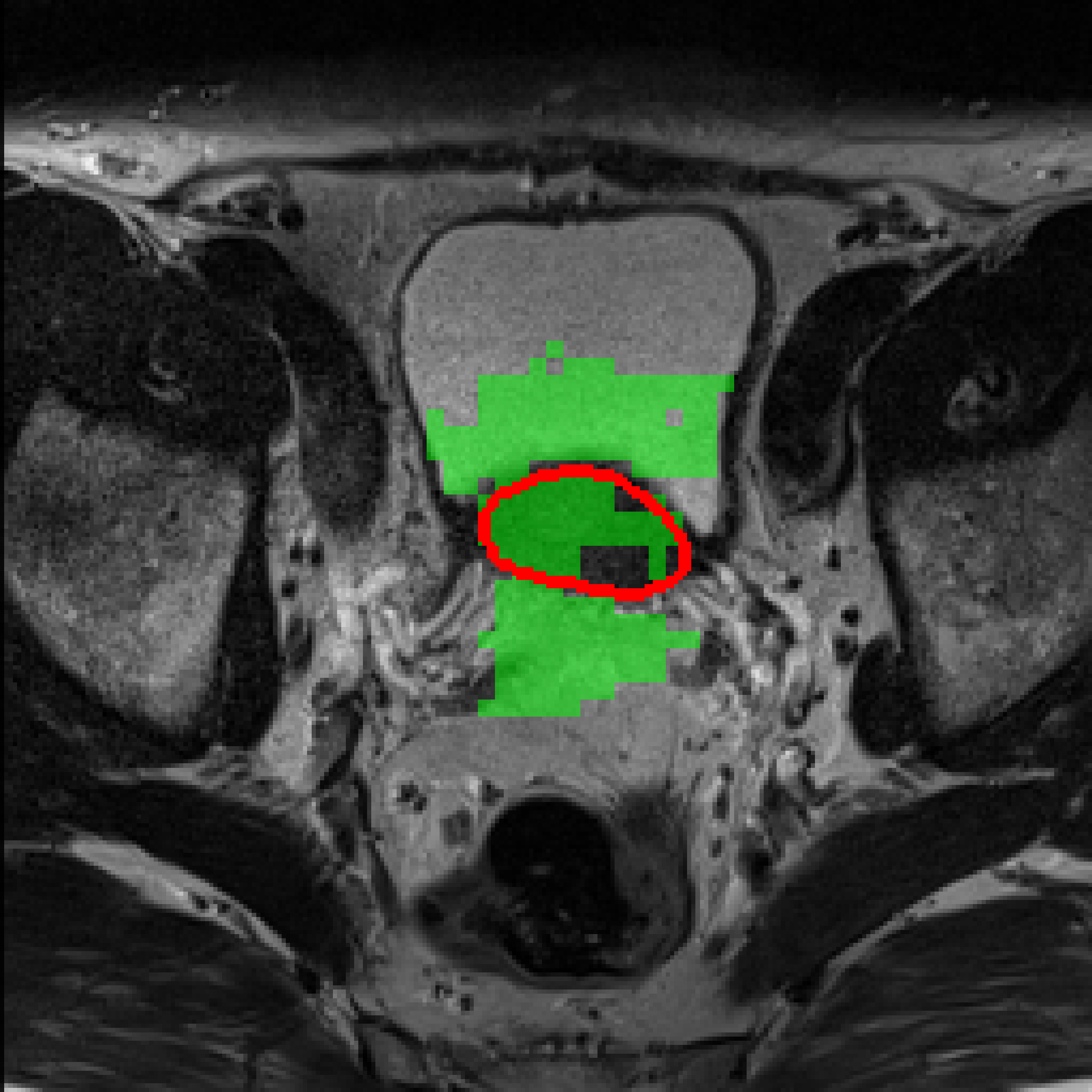} &
    \includegraphics[width=0.15\columnwidth]{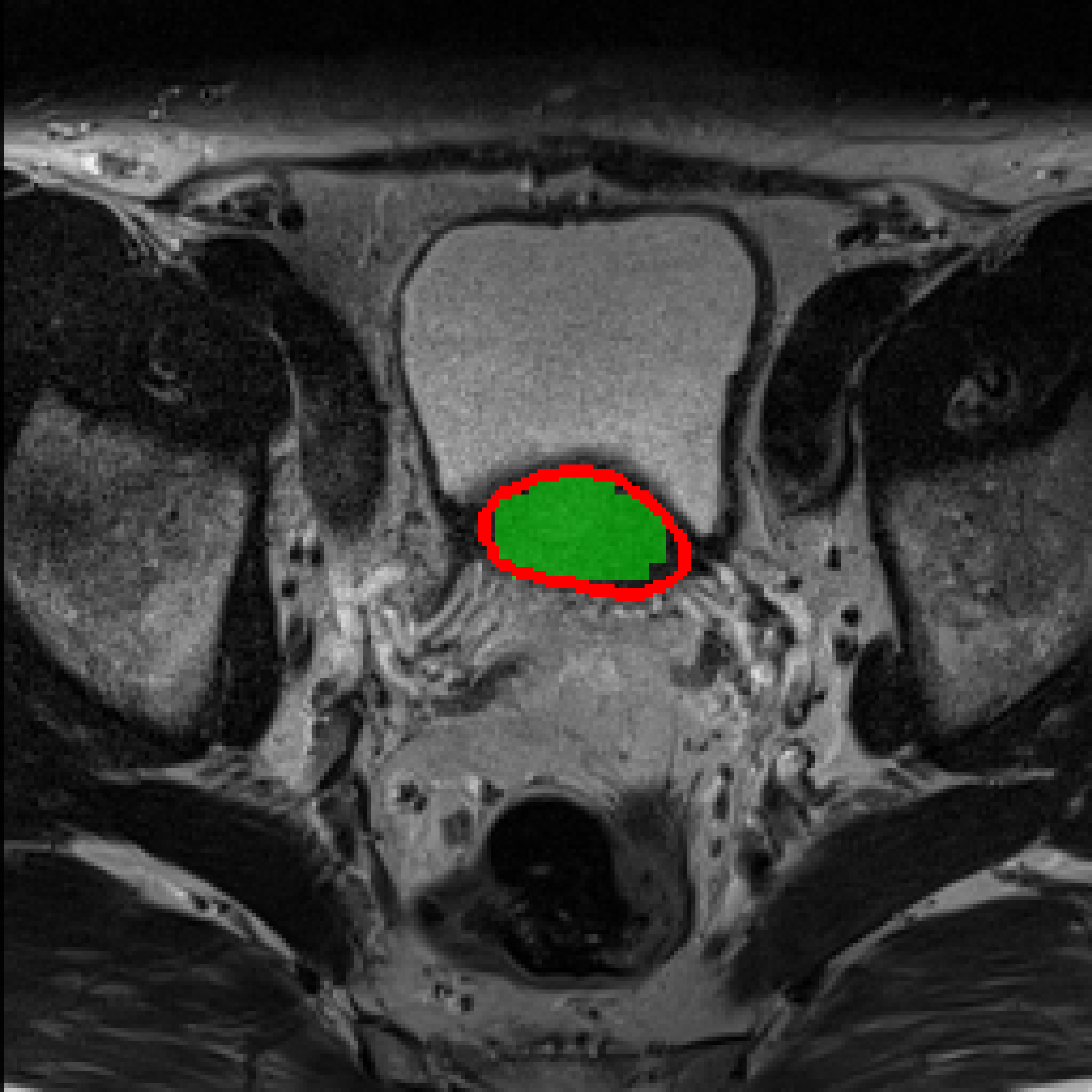} \\[2pt]

    \adjustbox{valign=c,rotate=90}{\tiny \textbf{10\% Promise}} &
    \includegraphics[width=0.15\columnwidth]{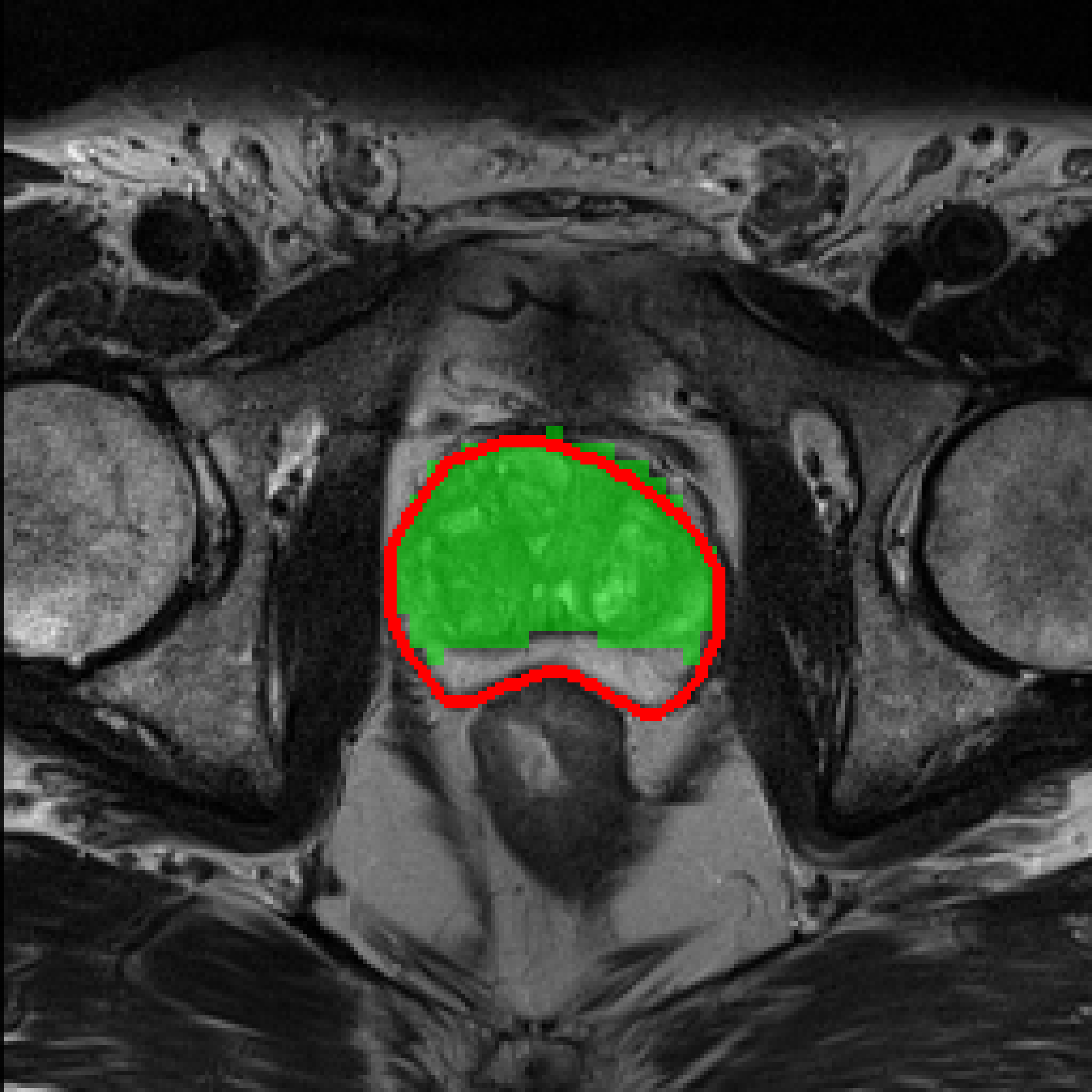} &
    \includegraphics[width=0.15\columnwidth]{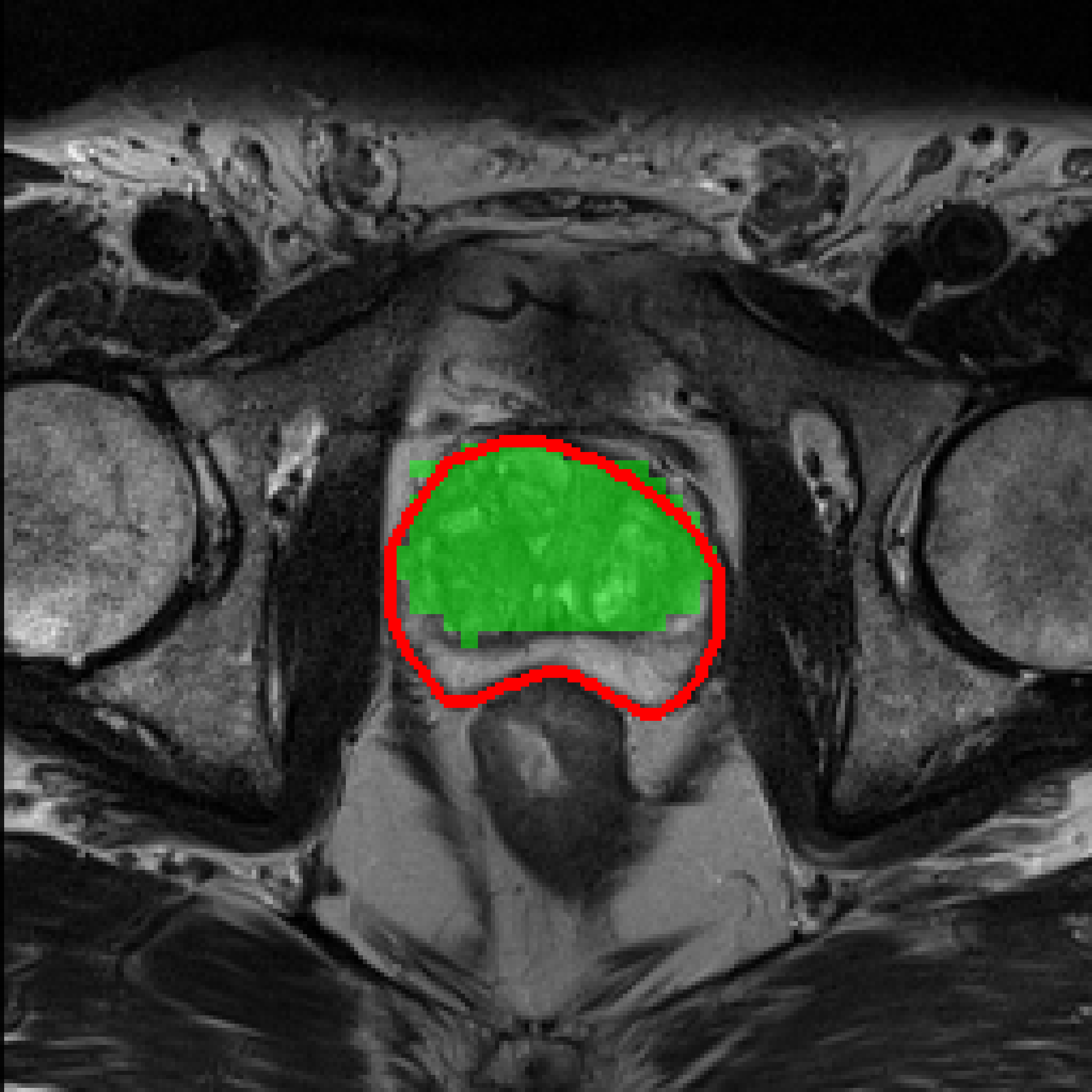} &
    \includegraphics[width=0.15\columnwidth]{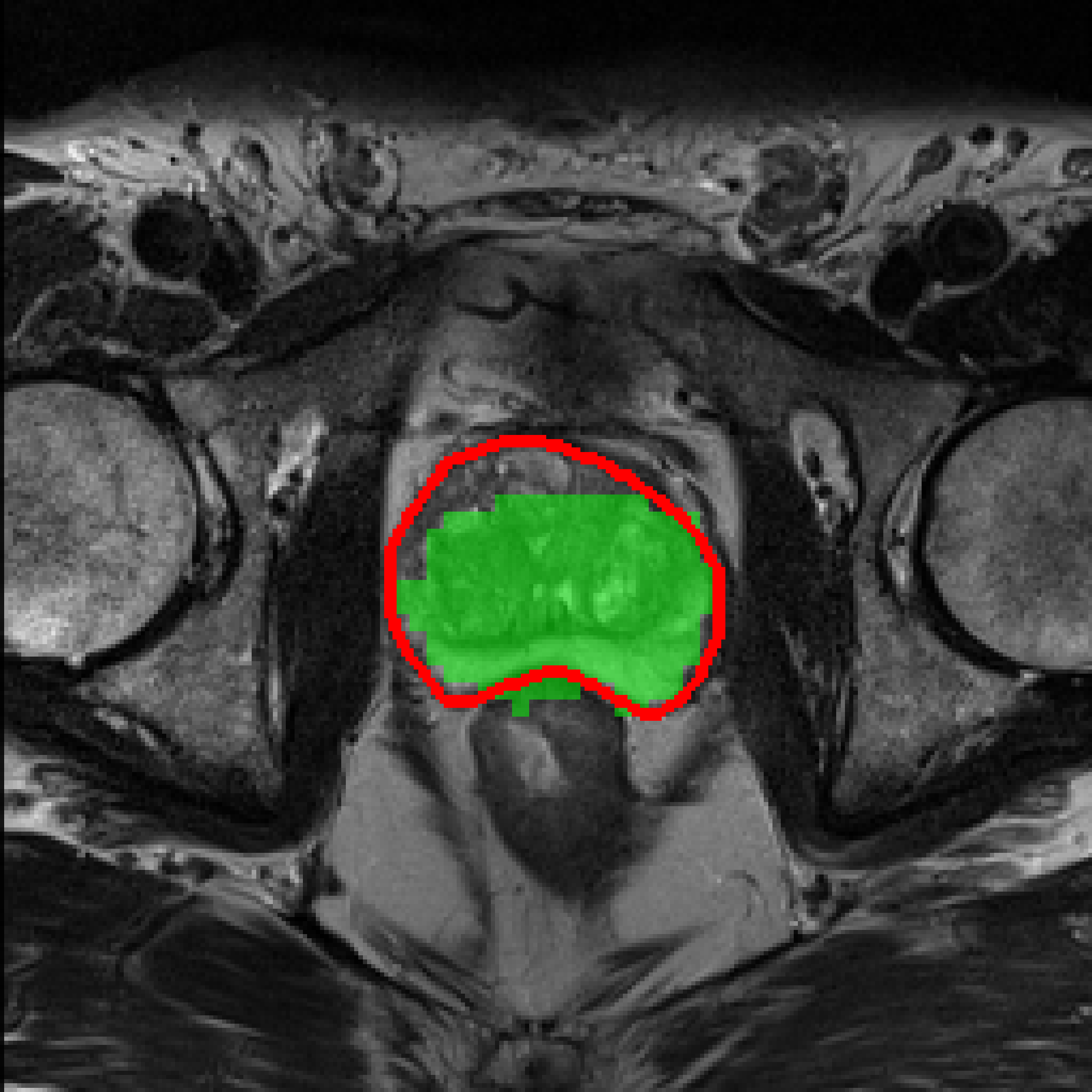} &
    \includegraphics[width=0.15\columnwidth]{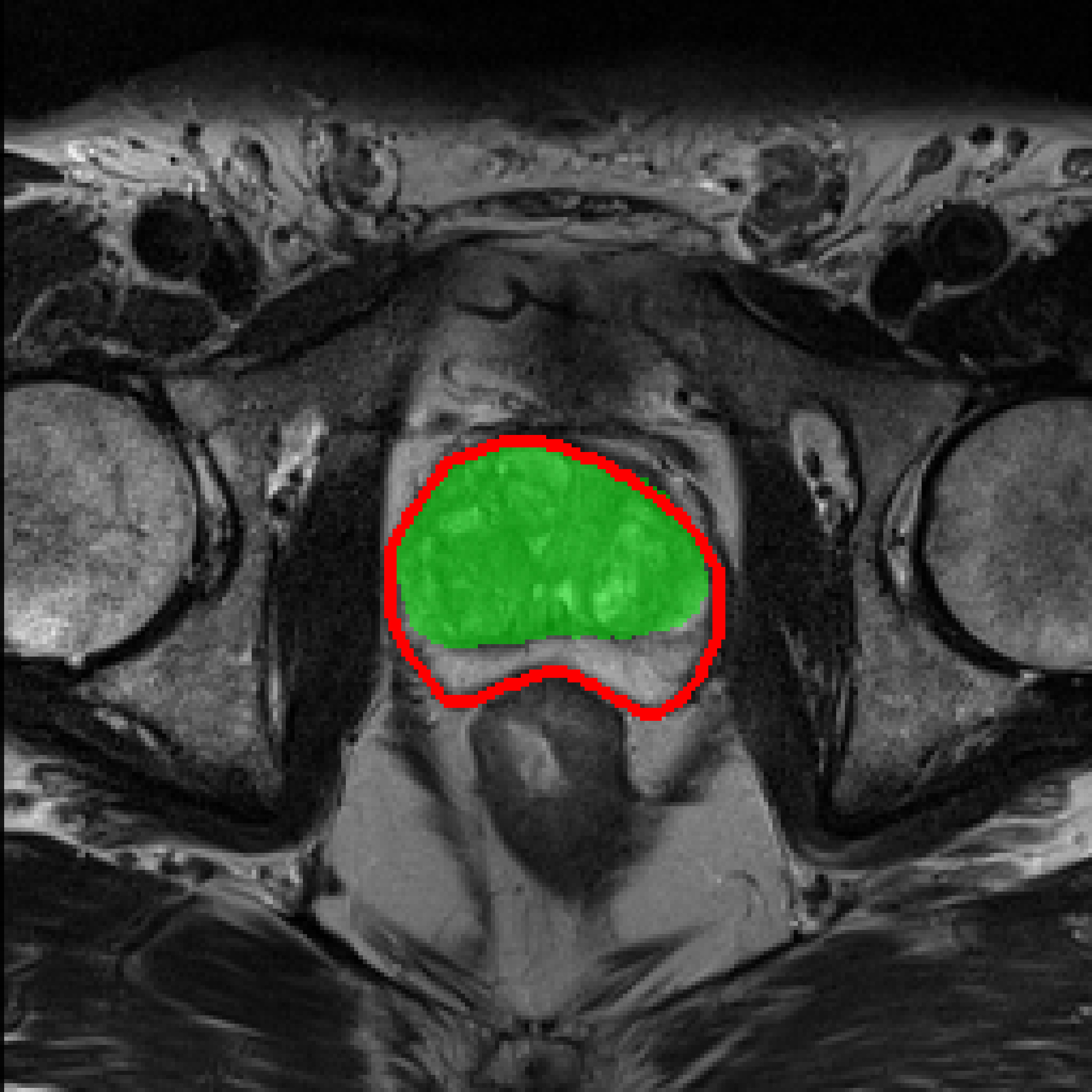} &
    \includegraphics[width=0.15\columnwidth]{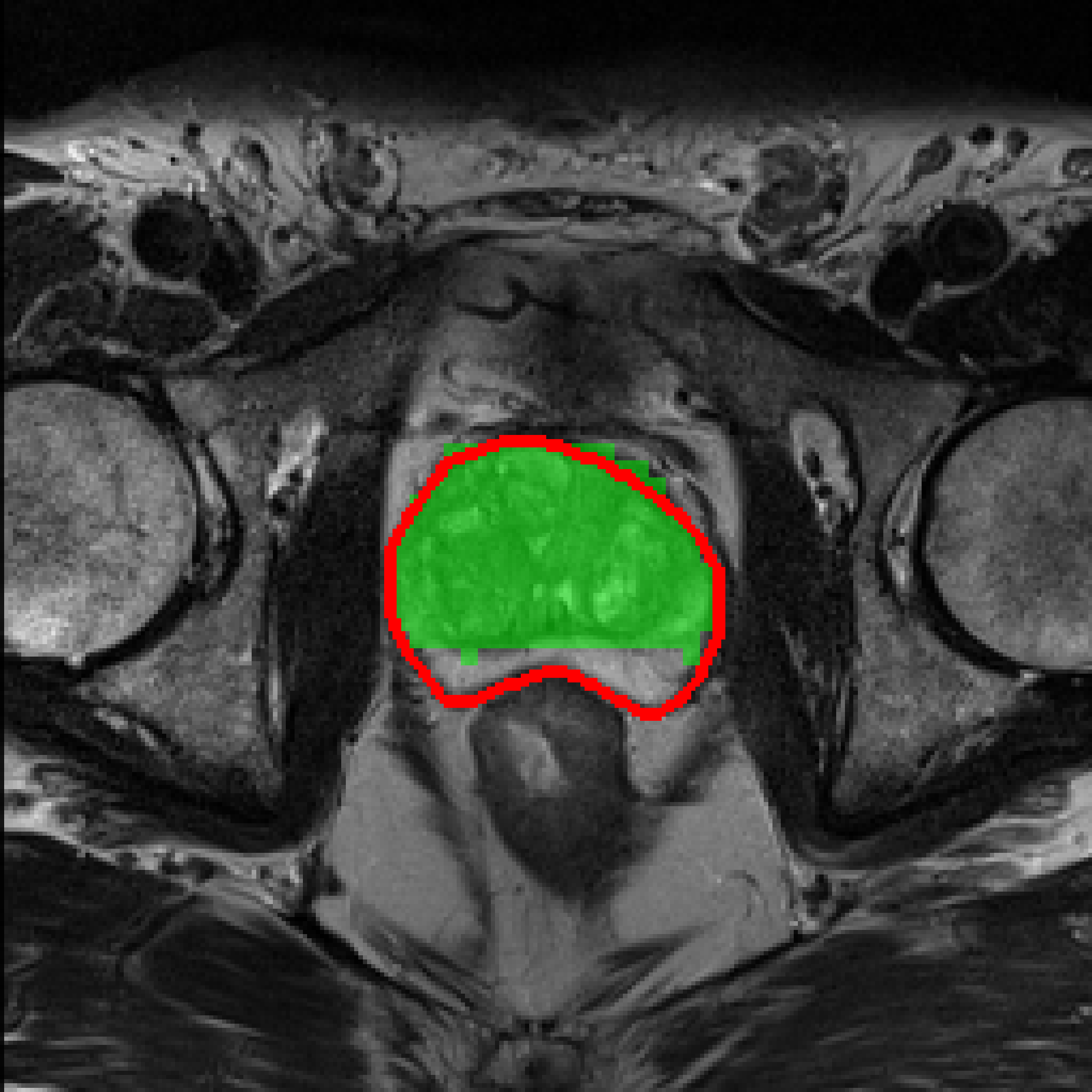} &
    \includegraphics[width=0.15\columnwidth]{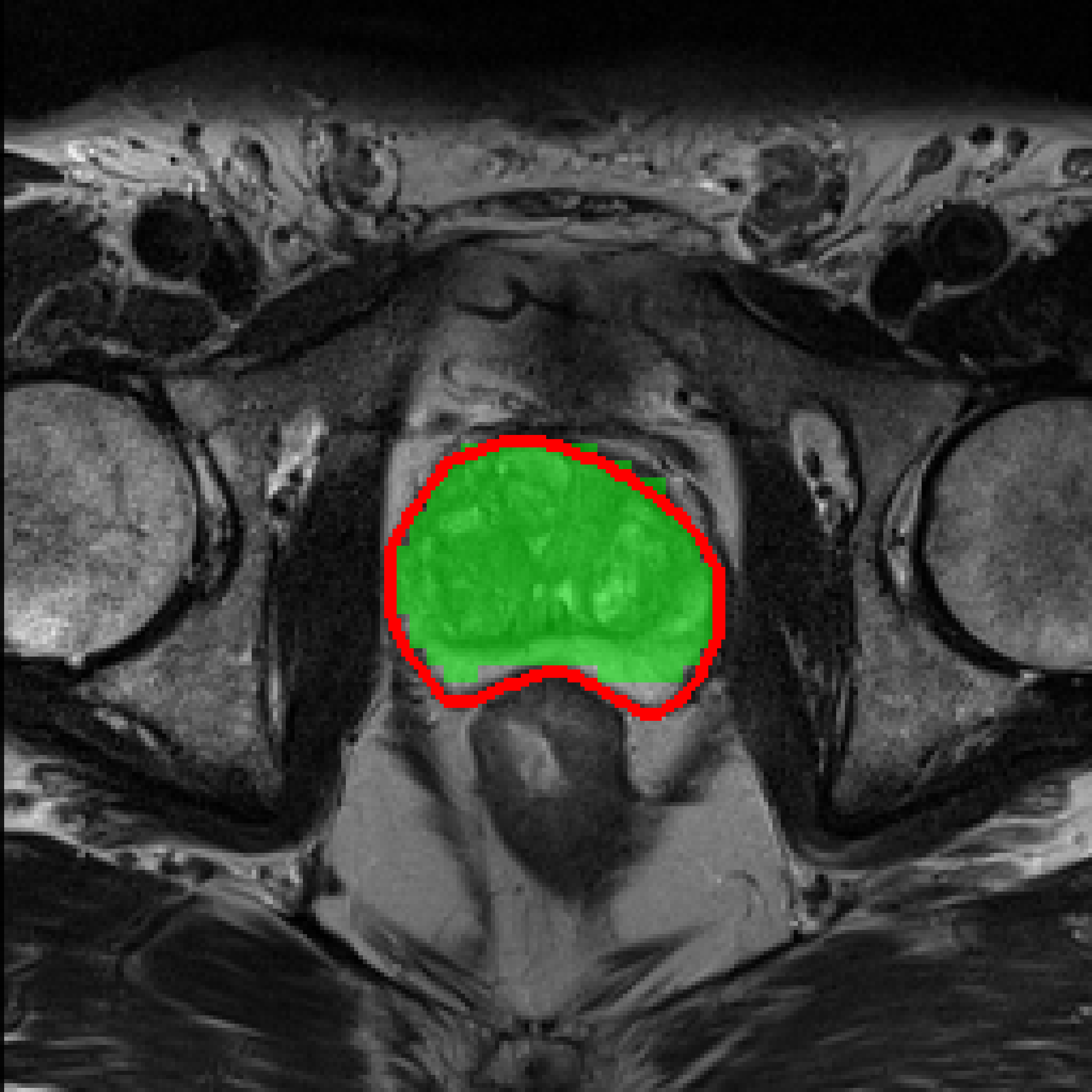} \\[2pt]

    \adjustbox{valign=c,rotate=90}{\tiny \textbf{5\% Colon}} &
    \includegraphics[width=0.15\columnwidth]{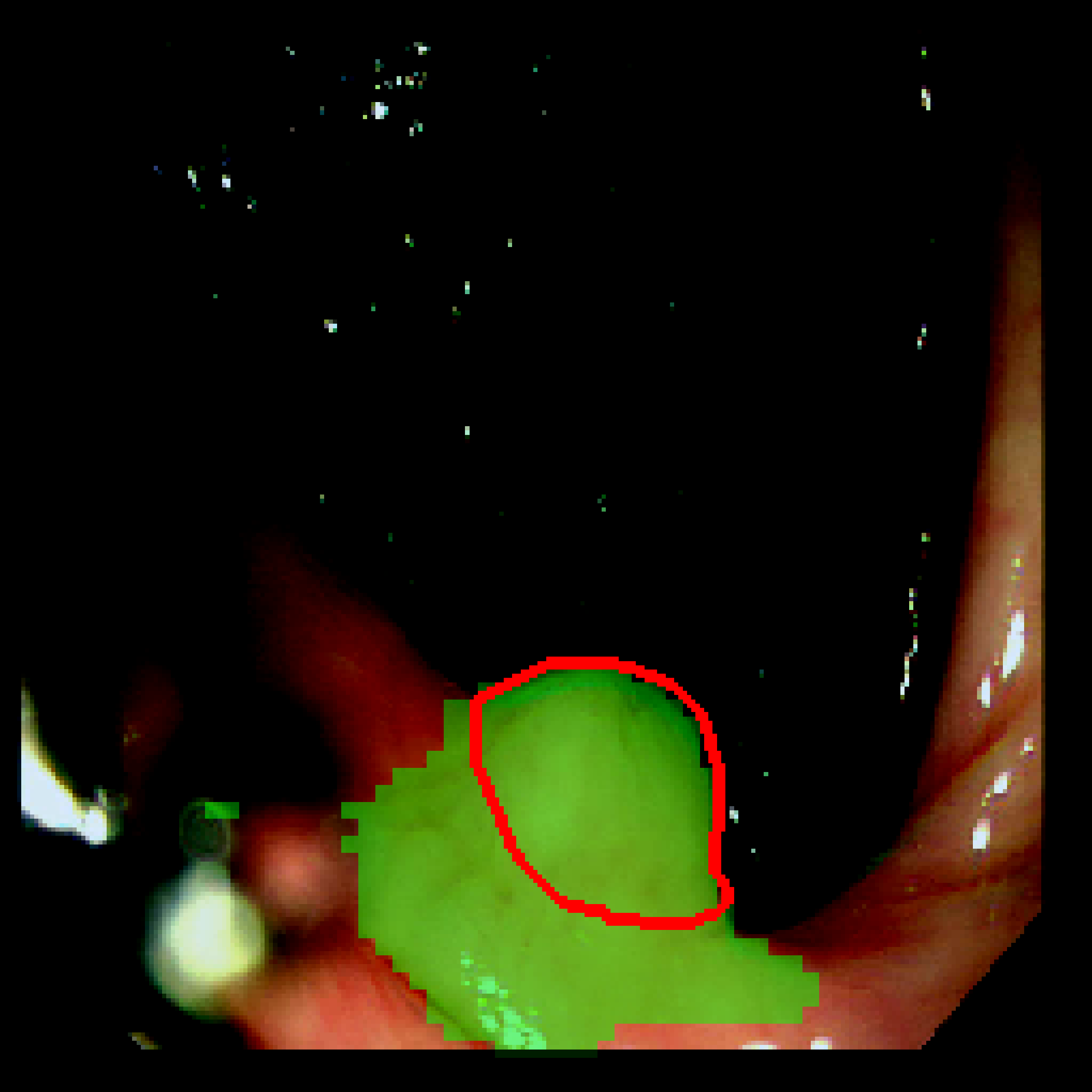} &
    \includegraphics[width=0.15\columnwidth]{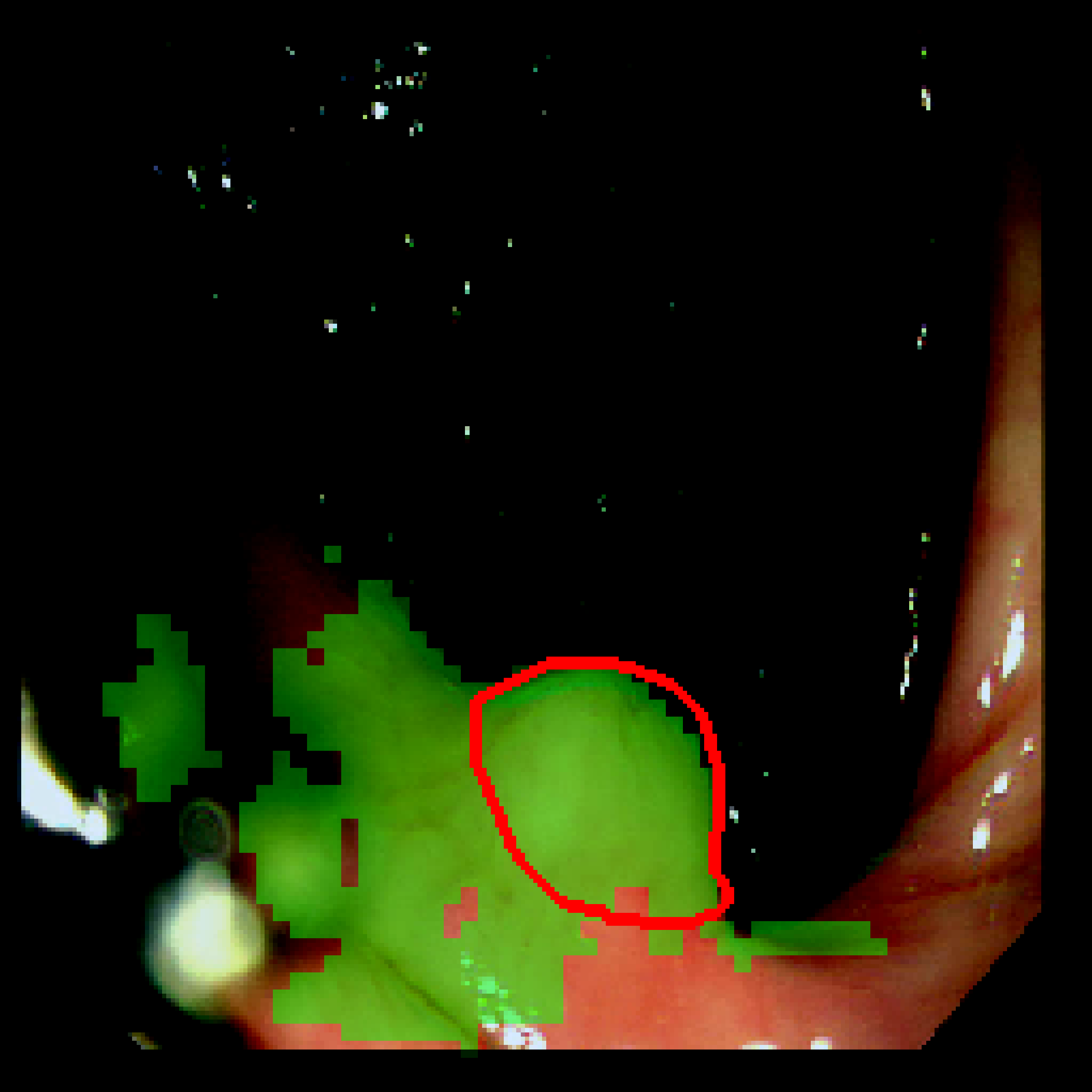} &
    \includegraphics[width=0.15\columnwidth]{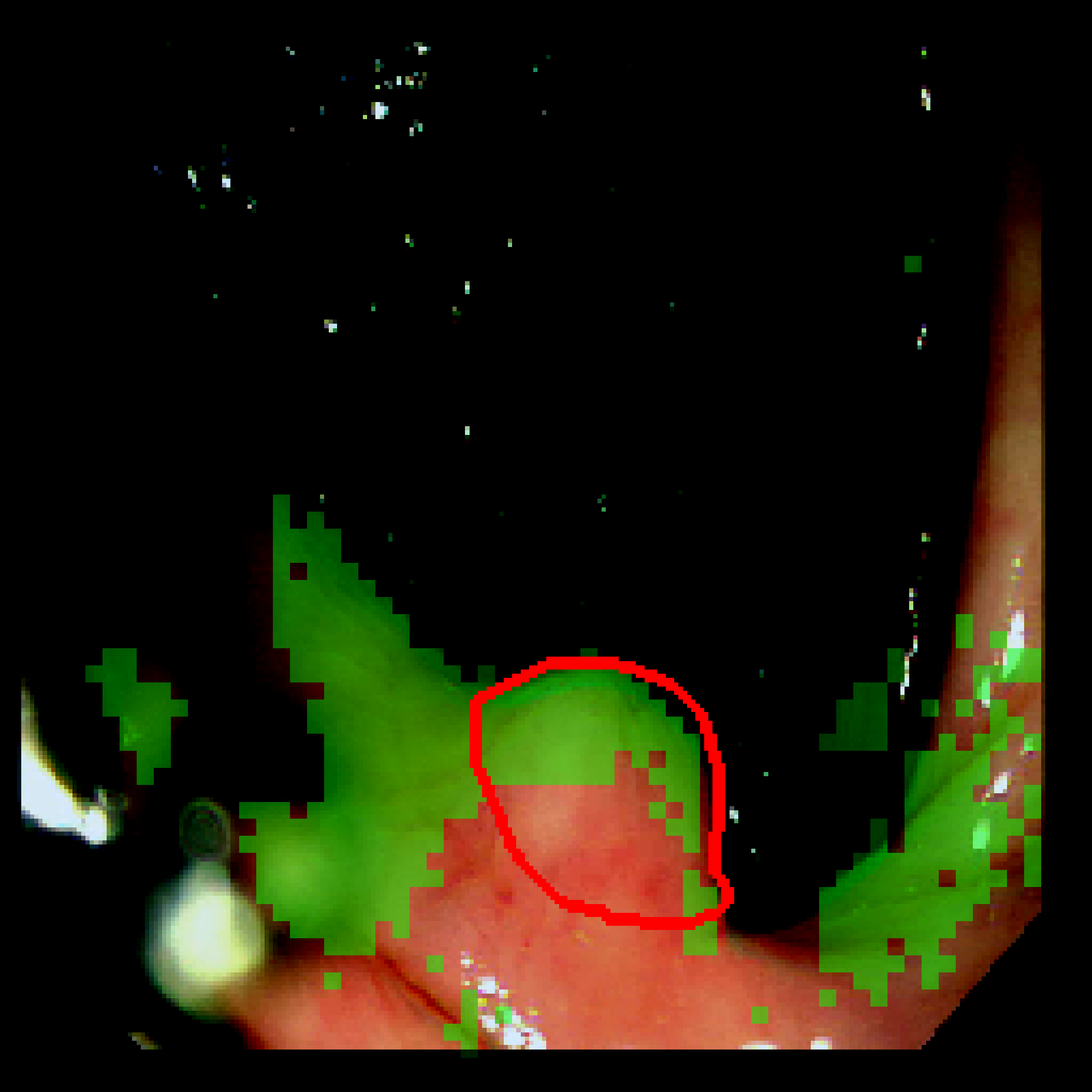} &
    \includegraphics[width=0.15\columnwidth]{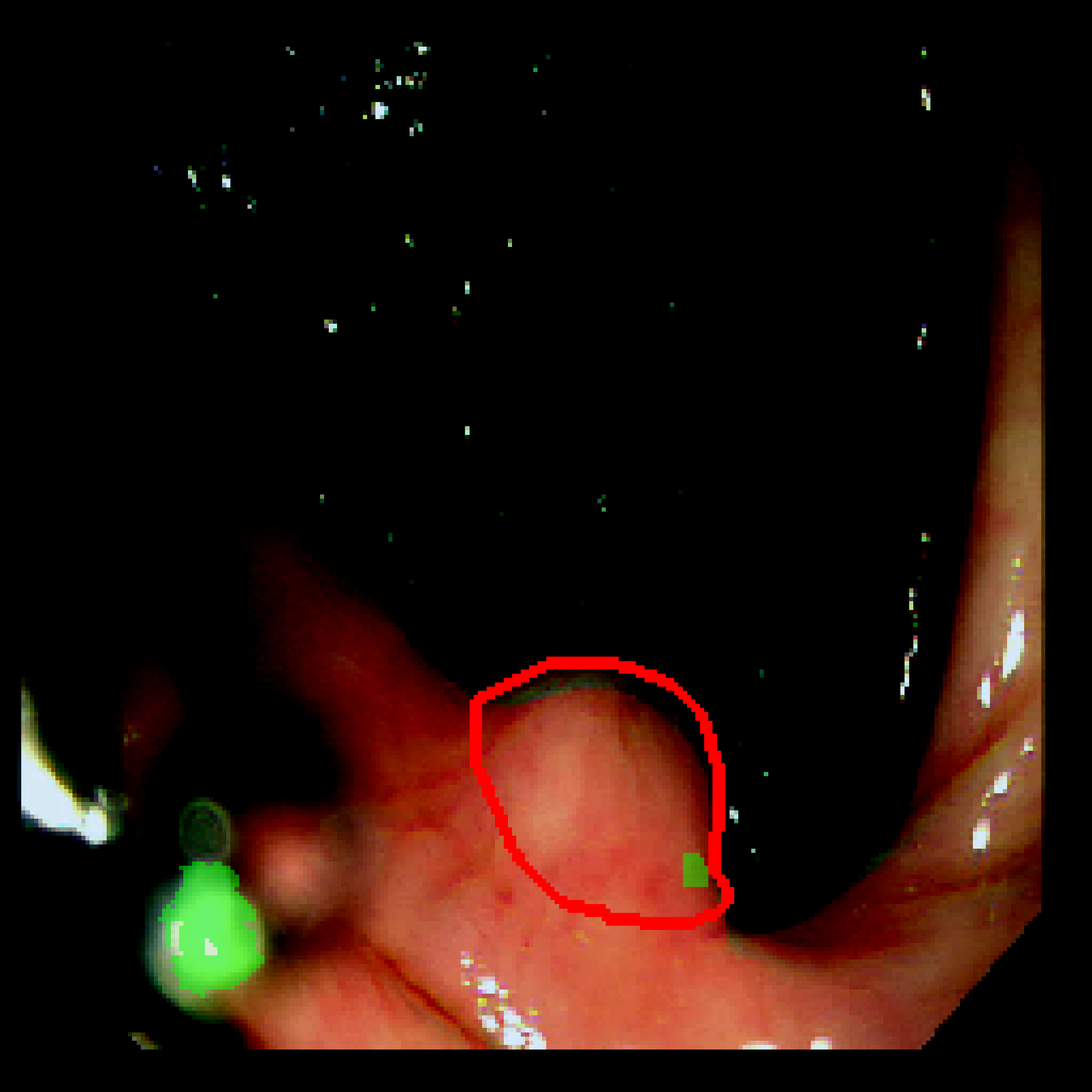} &
    \includegraphics[width=0.15\columnwidth]{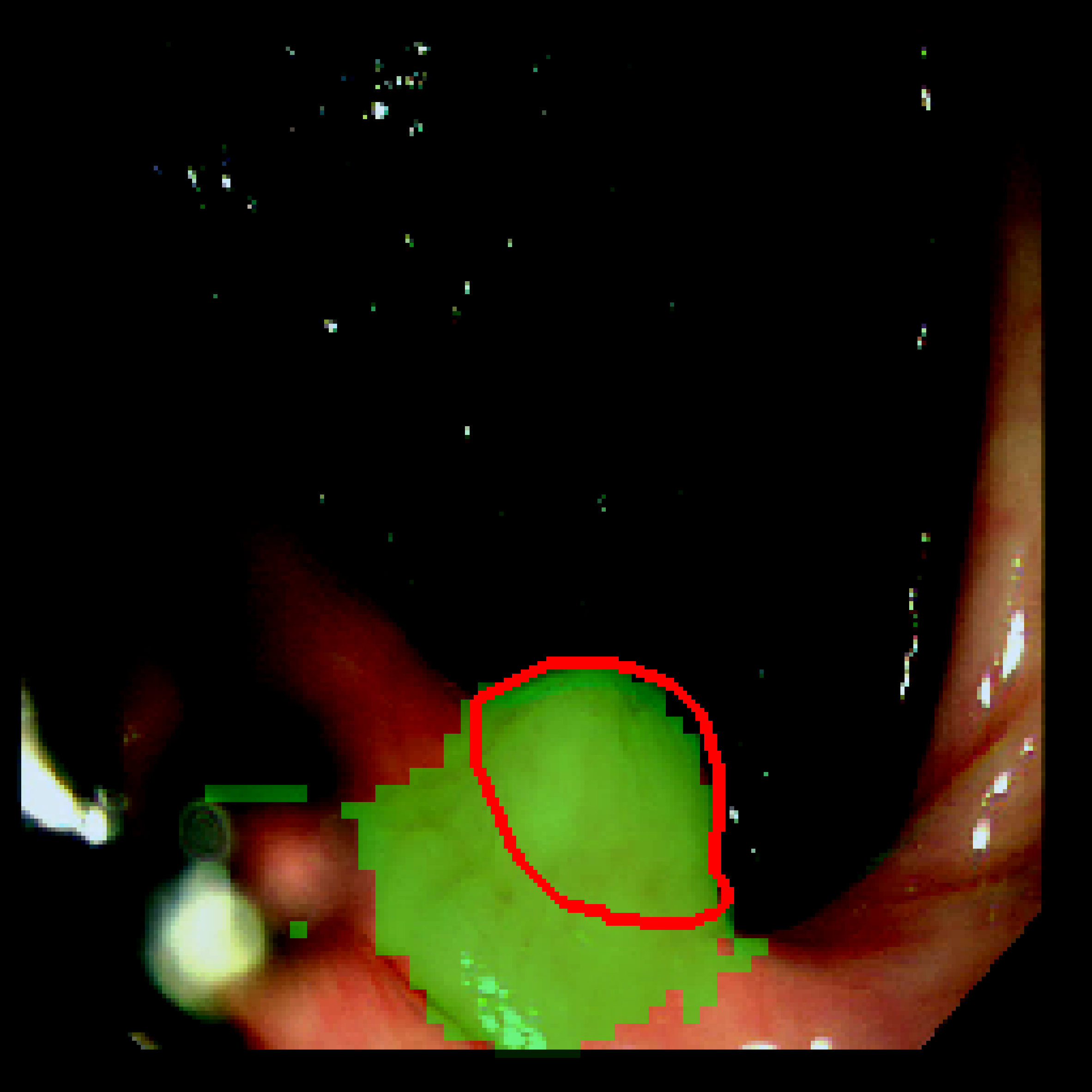} &
    \includegraphics[width=0.15\columnwidth]{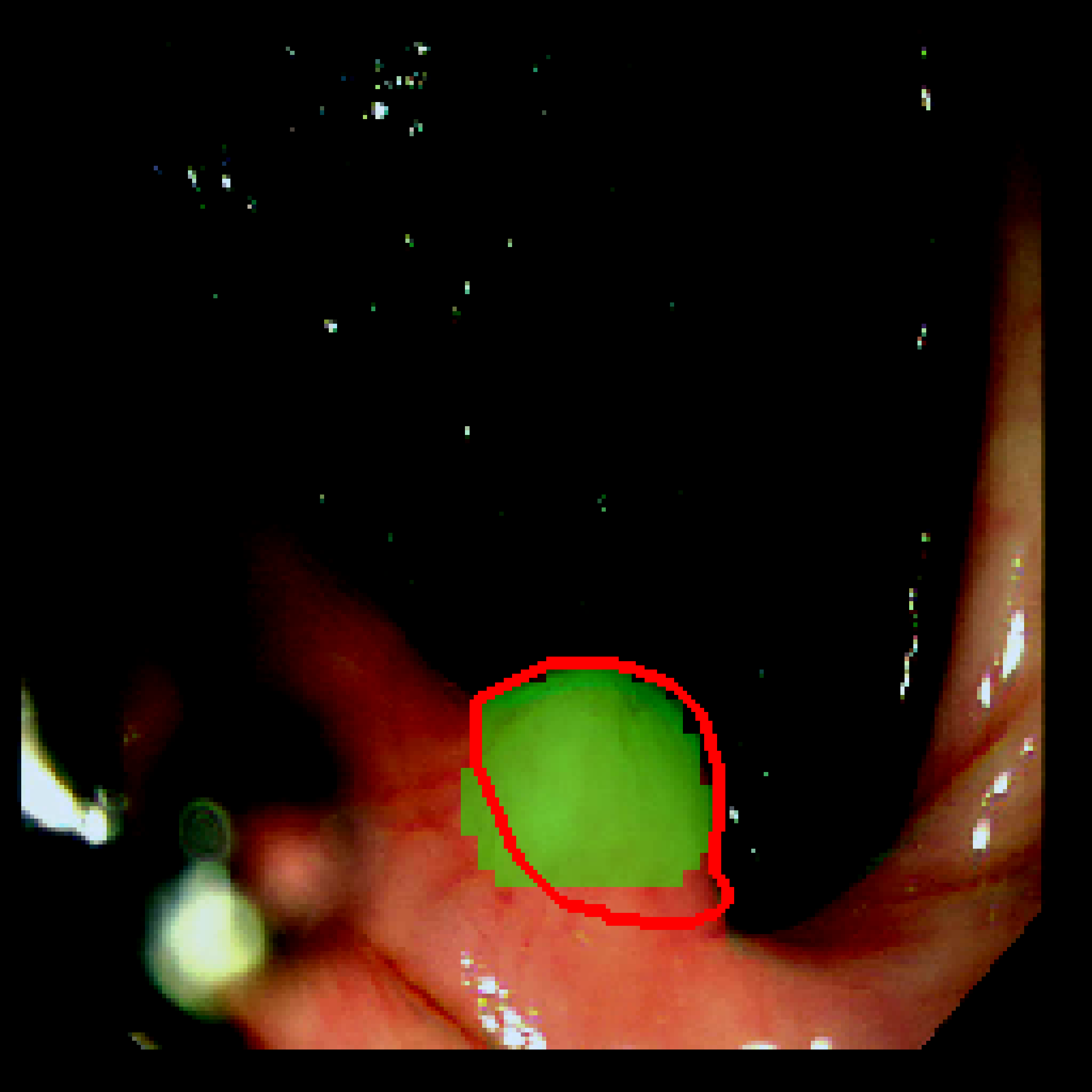} \\[2pt]

    \adjustbox{valign=c,rotate=90}{\tiny \textbf{10\% Colon}} &
    \includegraphics[width=0.15\columnwidth]{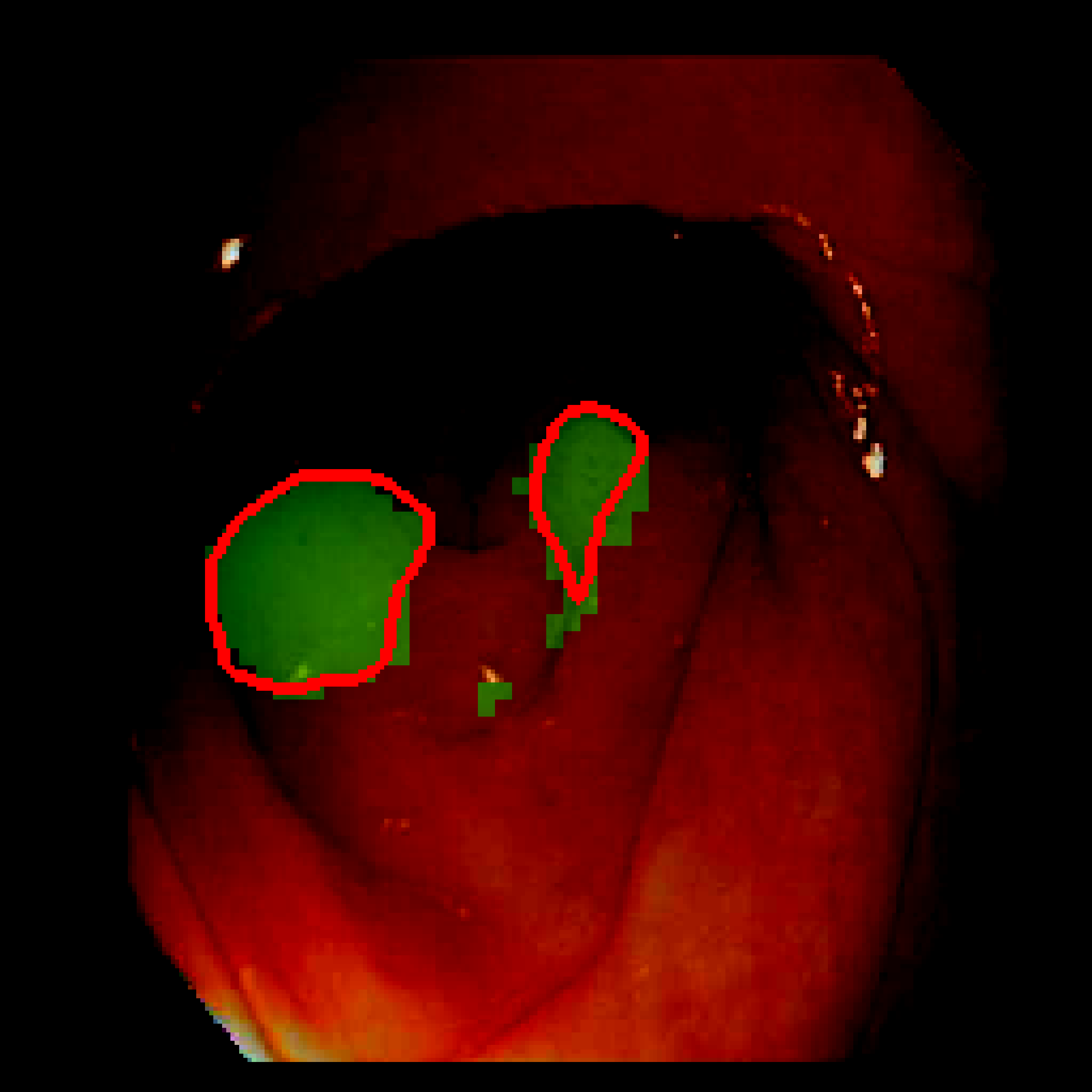} &
    \includegraphics[width=0.15\columnwidth]{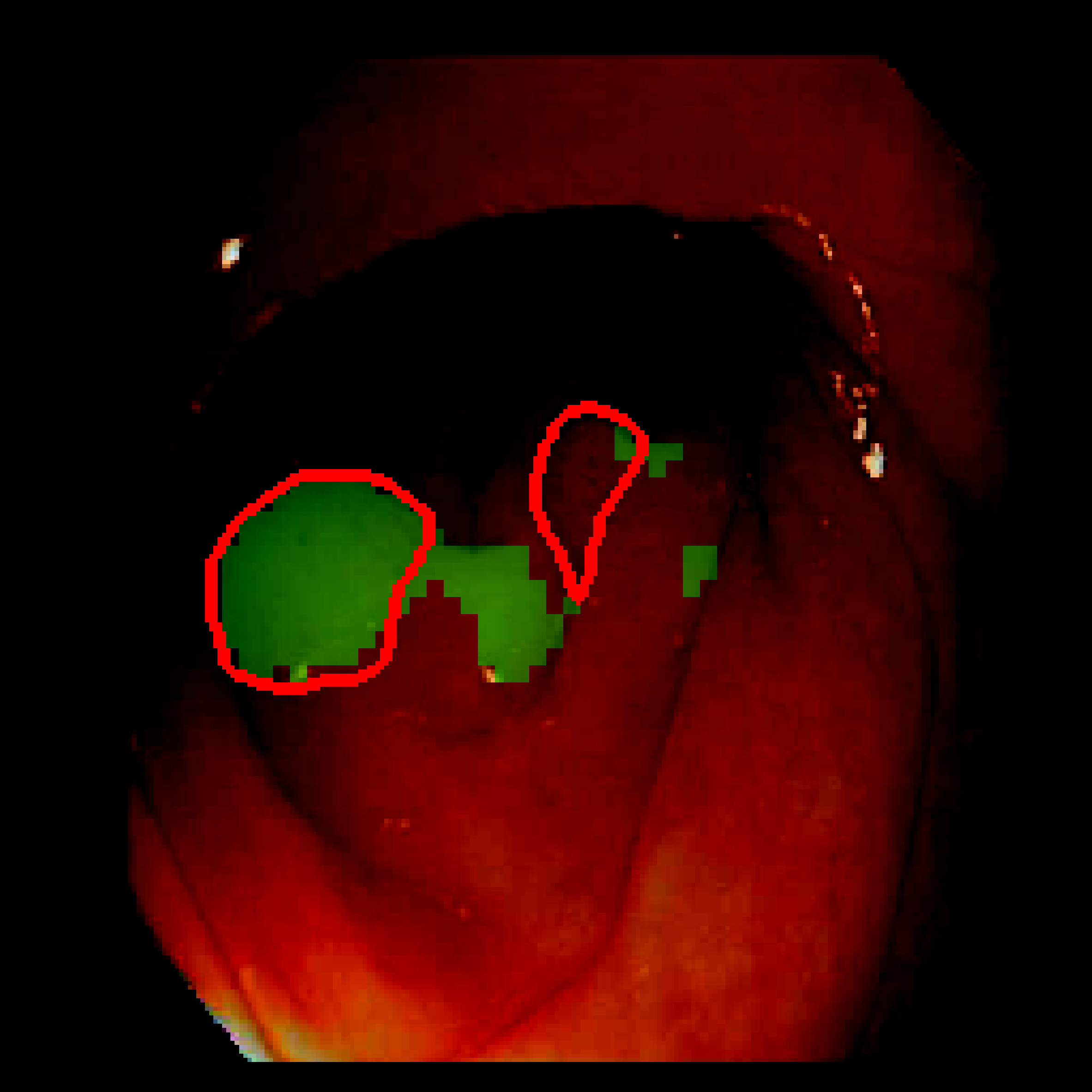} &
    \includegraphics[width=0.15\columnwidth]{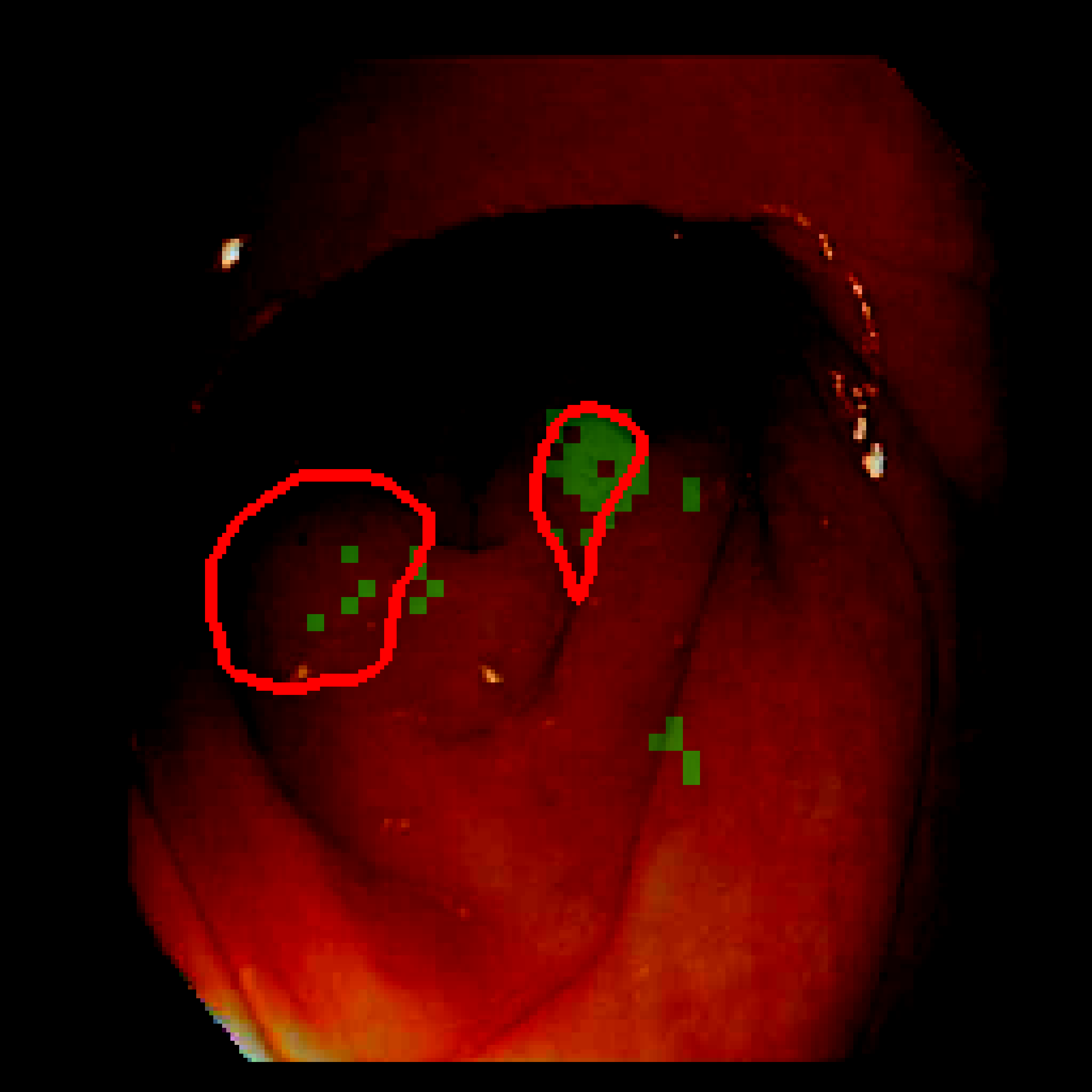} &
    \includegraphics[width=0.15\columnwidth]{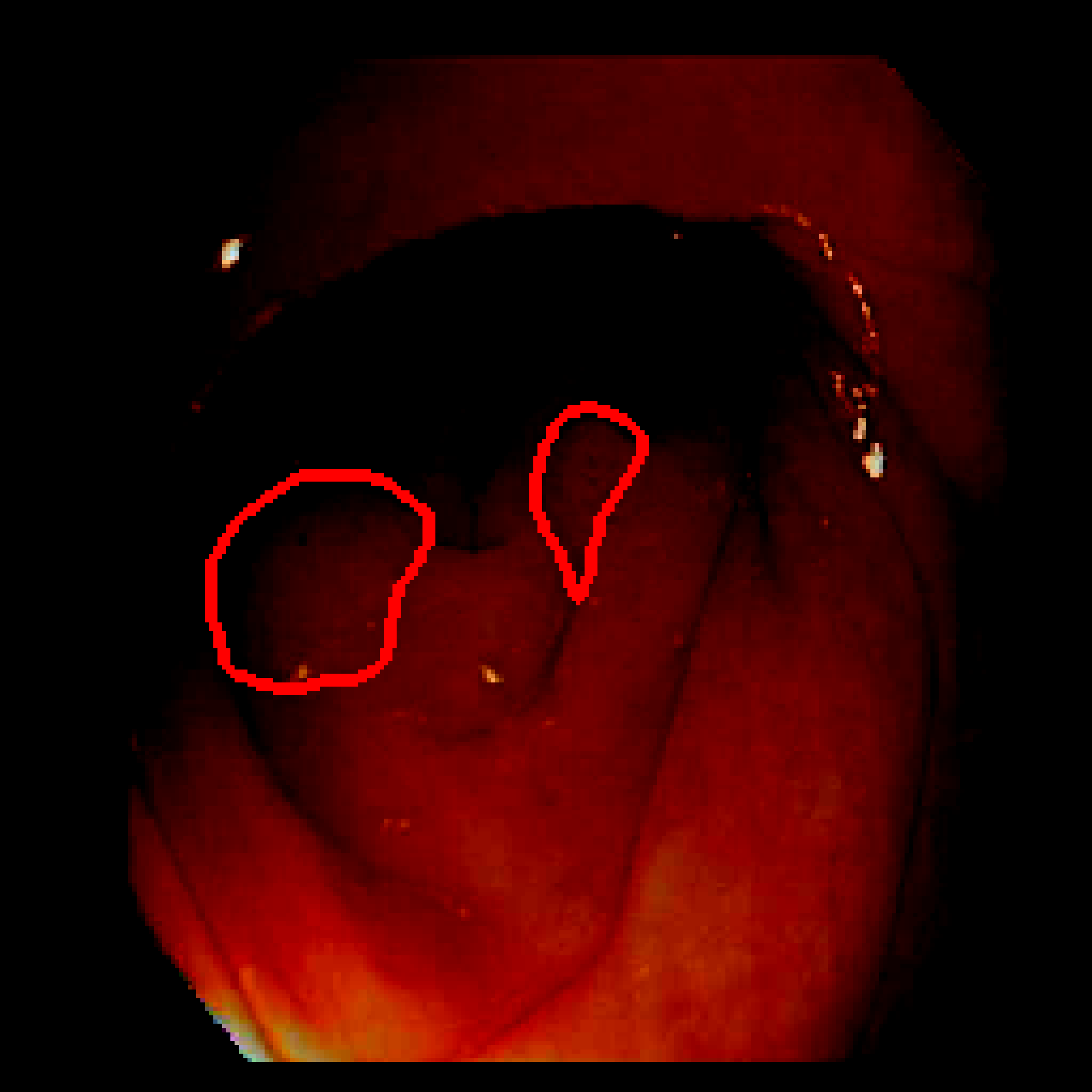} &
    \includegraphics[width=0.15\columnwidth]{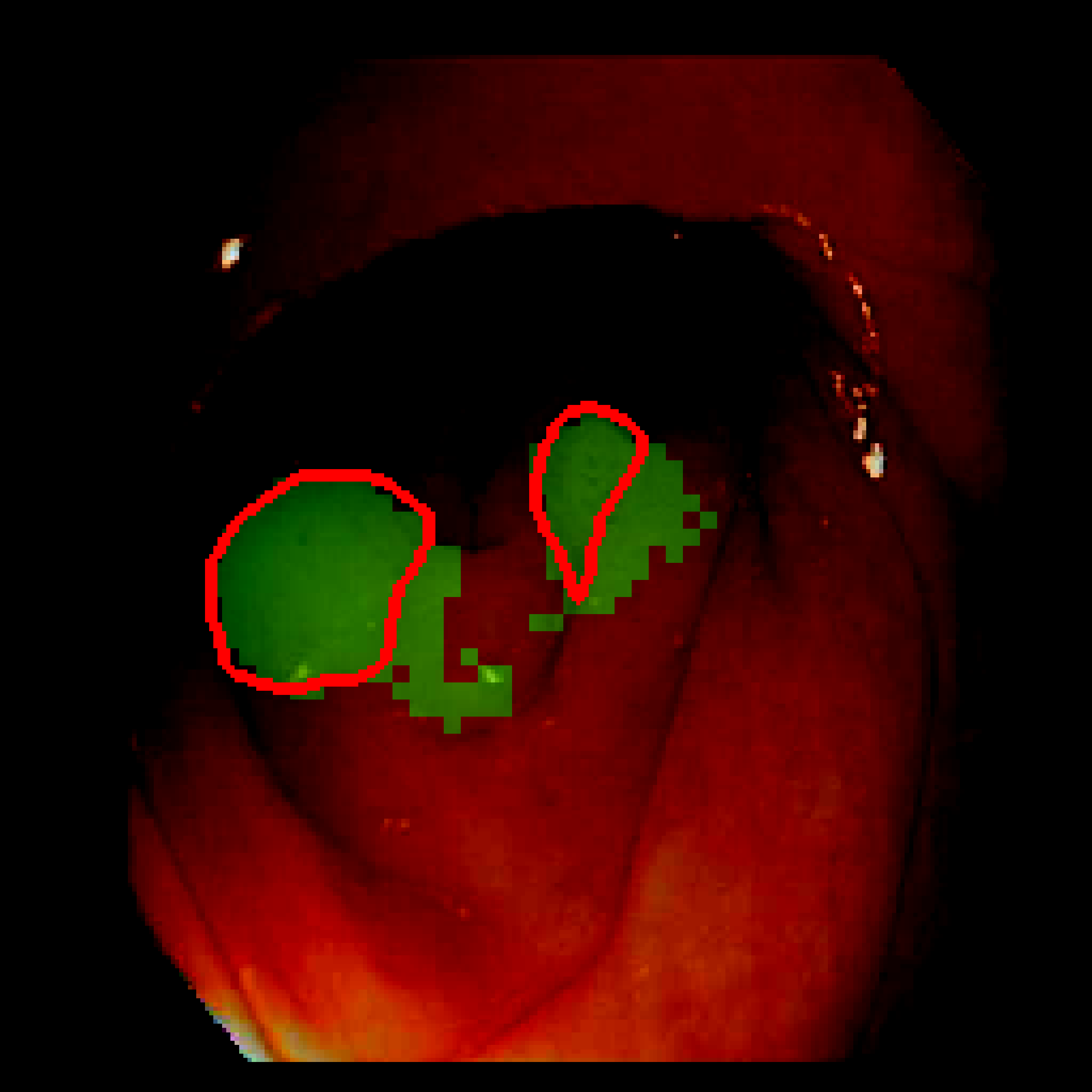} &
    \includegraphics[width=0.15\columnwidth]{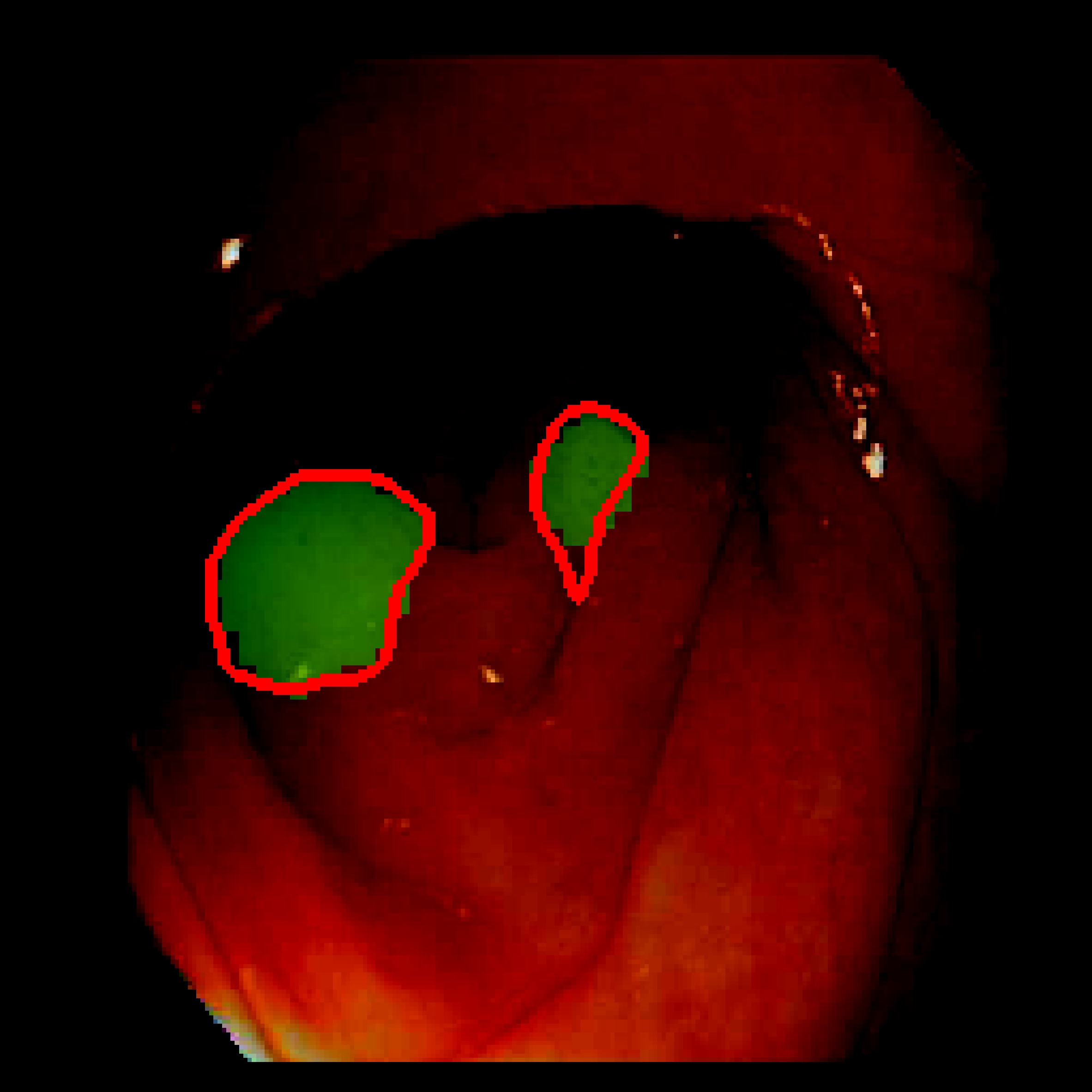}
\end{tabular}
\end{table}
\end{center}

For the qualitative results shown in Tab. \ref{tab:qual}, our method consistently produces more accurate and compact segmentation results for most cases in the PROMISE12 and COLON datasets, effectively capturing the prostate and polyp boundaries while suppressing background noise. In contrast, other approaches tend to over-segment the target regions, leading to the inclusion of irrelevant surrounding tissues. These results visually demonstrate the robustness of our method in handling diverse anatomical variations.

\subsection{Ablation Study}

To ascertain the effectiveness of each component in our framework, we performed ablation studies on both the specialist backbone selection and the use of a sigmoid ramp-up during unsupervised phase. As shown in Tab.~\ref{tab:ablation}, directly transferring supervision from UNet to SAM without ramp-up caused noisy unlabeled signals to dominate the early stage of SAM tuning, resulting in a 47\% drop in Dice score. Replacing the UNet specialist with UNet++ \cite{unetpp} or ResUNet++ \cite{resunetpp} yielded lower performance. Swin-UNet \cite{swinunet} performs as a less effective pseudo-label generator than CNN-based backbones due to the data-hungry nature of ViTs in semi-supervised settings.

\begin{center}
\vspace{-5mm}
\begin{table}[h!]
\centering
\caption{Ablation study on sigmoid ramp-up and specialist backbones with 5\% labeled data on PROMISE12 dataset.\\
}
\label{tab:ablation}
\renewcommand{\arraystretch}{1.1}
\setlength{\tabcolsep}{7pt}
\resizebox{1\linewidth}{!}{
\begin{tabular}{lccccc}
\hline
\textbf{Backbone} & \textbf{Ramp-up} & \textbf{Dice} & \textbf{IoU} & \textbf{HD95} & \textbf{ASD} \\ 
\hline
UNet \cite{unet}  &           & 36.37 & 29.46 & 47.26 & 49.07 \\ 
SwinUNet \cite{swinunet} & $\checkmark$ & 70.27 & 56.44 &  4.26 &  8.12 \\ 
ResUNet++ \cite{resunetpp} & $\checkmark$ & 78.20 & 66.32 &  4.09 &  5.17 \\ 
UNet++ \cite{unetpp}  & $\checkmark$ & 80.14 & 68.87 &  4.05 &  4.56 \\ 
\rowcolor{red!7} 
UNet \cite{unet} & $\checkmark$ & \textbf{83.64} & \textbf{73.87} & \textbf{3.98} & \textbf{3.79} \\
\hline
\end{tabular}
}
\end{table}
\vspace{-10mm}
\end{center}

\section{Conclusion}
We presented \textit{SC-SAM}, a specialist–generalist framework that enables U-Net to guide a PEFT SAM in leveraging unlabeled medical images. Through a stable bidirectional co-training loop, U-Net provides structural cues while SAM offers refined semantic regularization, allowing SAM to benefit directly from semi-supervised learning. Experiments on prostate MRI and polyp segmentation show that \textit{SC-SAM} surpasses recent semi-supervised SAM variants and even strong medical foundation models. These results demonstrate that combining a conventional specialist with a powerful generalist offers an effective and label-efficient strategy for medical image segmentation.

\section{Acknowledgment}
This work was supported in part by U.S. NIH grant R35GM158094. We thank AI VIETNAM for supporting us with GPUs to conduct experiments.





\bibliographystyle{IEEEbib}
\bibliography{strings,refs}

\end{document}